\documentclass[11pt, a4paper, logo]{googlecloud}

\pdftrailerid{redacted}

\makeatletter
\renewcommand\bibentry[1]{\nocitep{#1}{\frenchspacing\@nameuse{BR@r@#1\@extra@b@citeb}}}
\makeatother

\definecolor{pastelgreen}{rgb}{0.88, 1.0, 0.88}
\definecolor{pastelred}{rgb}{1.0, 0.88, 0.88}
\definecolor{ARCBlue}{RGB}{20, 50, 100}

\usepackage{dsfont}

\usepackage[authoryear, compress, round]{natbib}

\usepackage[utf8]{inputenc} %

\usepackage{microtype}
\usepackage{graphicx}
\usepackage{subcaption}
\usepackage{booktabs} 

\usepackage{hyperref}
\usepackage{wrapfig}

\usepackage{amsmath}
\usepackage{amssymb}
\usepackage{mathtools}
\usepackage{amsthm}

\usepackage[capitalize,noabbrev]{cleveref}

\theoremstyle{plain}

\theoremstyle{definition}

\theoremstyle{remark}

\usepackage[T1]{fontenc}

\usepackage{microtype}

\usepackage{amssymb,bm,mathtools,color}

\usepackage[normalem]{ulem}
\usepackage{enumitem}

\usepackage{booktabs}
\usepackage{multirow}
\usepackage{tabularx}
\usepackage{multicol}
\usepackage{multirow}
\usepackage{colortbl}
\usepackage{hhline}
\usepackage{stfloats}

\newcolumntype{Y}{>{\centering\arraybackslash}X}

\usepackage{xcolor}
\definecolor{red}{rgb}{0.74,0.08,0.10}
\definecolor{green}{rgb}{0.26,0.49,0.18}
\definecolor{blue}{rgb}{0.22,0.53,0.75}

\usepackage{colortbl}
\definecolor{Gray}{gray}{0.9}
\definecolor{LightCyan}{rgb}{0.75,1,1}

\makeatletter
\newcommand\notsotiny{\@setfontsize\notsotiny\@viiipt\@ixpt}
\makeatother

\newcommand{\ie}{\emph{i.e.}, }
\newcommand{\eg}{\emph{e.g.}, }

\newcommand{\makename}[3][s]{%
  \expandafter\newcommand\csname #2\endcsname{#3\xspace}%
  \expandafter\newcommand\csname #2s\endcsname{#3#1\xspace}%
}

\makename{logitLens}{logit lens}
\makename{methodNameAbbr}{\textsc{acute}}
\makename{methodName}{activation-based confidence, utility, and trust estimation}
\makename{methodNameCaps}{Activation-based Confidence, Utility, and Trust Estimation}
\makename{metricName}{expected utility renormalized by the oracle}
\makename{metricNameFirstCaps}{Expected utility renormalized by the oracle}
\makename{metricNameCaps}{Expected Utility Renormalized by the Oracle}
\makename{metricNameAbbr}{\textsc{euro}}
\makename{euroauc}{\textsc{auc-euro}}
\makename{voleuro}{\texttt{Vol}$_{\metricNameAbbr}$}
\makename{uct}{$U_{ct}$}
\makename{uca}{$U_{ca}$}
\makename{normuct}{$u_{ct}$}
\makename{normuca}{$u_{ca}$}
\makename{rtp}{$R_{tp}$}
\makename{rtn}{$R_{tn}$}
\makename{rfp}{$R_{fp}$}
\makename{rfn}{$R_{fn}$}

\newcommand{\kernelReg}{NWKR}

\newcommand{\hiddenStateAtLayer}[1]{H^{(#1)}}
\newcommand{\hiddenTokenAtLayer}[2]{{\bf h}^{(#1)}_{#2}}
\newcommand{\meanPoolHiddenStateAtLayer}[1]{{\bf {\bar h}}^{(#1)}}

\newcommand{\cosineFeatures}{{\bf x}_{\textsc{cosine}}}

\usepackage{amsmath,amsfonts,bm}

\def\eqref#1{equation~\ref{#1}}

\def\1{\bm{1}}

\DeclareMathAlphabet{\mathsfit}{\encodingdefault}{\sfdefault}{m}{sl}
\SetMathAlphabet{\mathsfit}{bold}{\encodingdefault}{\sfdefault}{bx}{n}

\newcommand{\eqdef}{\mathbin{\stackrel{\rm def}{=}}}

\usepackage[]{todonotes}

\usepackage[]{todonotes}  
\makeatletter
\newcommand*\iftodonotes{\if@todonotes@disabled\expandafter\@secondoftwo\else\expandafter\@firstoftwo\fi}  
\makeatother

\usepackage{listings}

\lstdefinestyle{simple_lst_style}{
    columns=flexible,
    keywordstyle=\color{red},
    numberstyle=\color{gray},
    stringstyle=\color{green},
    basicstyle=\ttfamily\small,
    identifierstyle=\color{black},
    commentstyle=\color{blue},
    breakatwhitespace=false,
    breaklines=true,
    captionpos=b,
    keepspaces=false,
    numbersep=5pt,
    showspaces=false,
    showstringspaces=false,
    showtabs=false,
    tabsize=2,
    frame=single,
}

\lstdefinelanguage{Prompt}{
    morekeywords={Human, Computer},
    sensitive=true,
    morestring=[b]",
}

\usepackage{booktabs}
\usepackage{multirow}
\usepackage{graphicx}

\usepackage{pifont}
\usepackage{amsmath}
\usepackage{amssymb}
\usepackage{ifthen}
\usepackage{dsfont}
\usepackage{bold-extra}
\usepackage{courier}

\usepackage{listings}
\usepackage{xcolor}
\usepackage{ltablex}
\usepackage{longtable}
\usepackage{multicol}

\usepackage{tikz}
\usetikzlibrary{bayesnet}
\usepackage[linguistics]{forest}

\usepackage{caption}

\usepackage{cancel}

\usepackage{xspace}
\usepackage{comment}
\usepackage{color,soul}

\usepackage{float}

\usepackage[noabbrev,capitalize]{cleveref} 

\crefname{page}{page}{pages}
\crefname{footnote}{footnote}{footnotes}   
\crefname{equation}{equation}{equations}   

\creflabelformat{equation}{#2\textup{#1}#3}

\crefname{line}{line}{lines}               
\crefrangeformat{line}{lines #3#1#4--#5#2#6}
\crefname{lstlsting}{Listing}{Listings}   
\crefname{section}{\S}{\S\S}
\Crefname{section}{\S}{\S\S}    
\crefformat{section}{#2\S#1#3}  
\Crefformat{section}{#2\S#1#3}
\crefrangeformat{section}{\S\S#3#1#4--#5#2#6}
\Crefrangeformat{section}{\S\S#3#1#4--#5#2#6}
\crefmultiformat{section}{\S#2#1#3}{ and~\S#2#1#3}{, \S#2#1#3}{ and~\S#2#1#3}
\Crefmultiformat{section}{\S#2#1#3}{ and~\S#2#1#3}{, \S#2#1#3}{ and~\S#2#1#3}
\crefrangemultiformat{section}{\S\S#3#1#4--#5#2#6}{ and~\S\S#3#1#4--#5#2#6}{, \S\S#3#1#4--#5#2#6}{ and~\S\S#3#1#4--#5#2#6}
\Crefrangemultiformat{section}{\S\S#3#1#4--#5#2#6}{ and~\S\S#3#1#4--#5#2#6}{, \S\S#3#1#4--#5#2#6}{ and~\S\S#3#1#4--#5#2#6}

\title{The \methodNameAbbr Protocol: Operationalizing Language Model Activations for Better Calibration, Utility, and Trust}
\correspondingauthor{Nishant Subramani \href{mailto:nishant2@cs.cmu.edu}{<nishant2@cs.cmu.edu>}, Hamid Palangi\href{mailto:hamidpalangi@google.com}{<hamidpalangi@google.com>}}

\renewcommand{\today}{}

\author[1,2]{\fontsize{10.0pt}{10.0pt}\selectfont Nishant Subramani}
\author[2]{\fontsize{10.0pt}{10.0pt}\selectfont Palash Goyal}
\author[2]{\fontsize{10.0pt}{10.0pt}\selectfont Yiwen Song}
\author[2]{\fontsize{10.0pt}{10.0pt}\selectfont Mani Malek}
\author[3]{\fontsize{10.0pt}{10.0pt}\selectfont Yuan Xue}
\author[2]{\fontsize{10.0pt}{10.0pt}\selectfont Tomas Pfister}
\author[2]{\fontsize{10.0pt}{10.0pt}\selectfont Hamid Palangi}

\affil[1]{\fontsize{9.0pt}{9.0pt}\selectfont Carnegie Mellon University}
\affil[2]{\fontsize{9.0pt}{9.0pt}\selectfont Google}
\affil[3]{\fontsize{9.0pt}{9.0pt}\selectfont Scale AI}

\begin{abstract}
As language models improve and become increasingly deployed to solve a variety of tasks, trustworthiness becomes essential.
Calibration is a good proxy for trust: well-calibrated confidence estimates help inform the risk versus reward tradeoff when trusting a specific model output.
Unfortunately, even as models improve, they remain poorly calibrated, often biasing towards overconfidence.
Additionally, calibration can be gamed: a policy that always predicts the base rate is perfectly calibrated, but completely uninformative.
To resolve this, we develop a new metric, \metricName (\metricNameAbbr), that balances calibration and informativeness.
We also propose a general-purpose \textbf{a}ctivation-based \textbf{c}onfidence, \textbf{u}tility, and \textbf{t}rust \textbf{e}stimation protocol (\methodNameAbbr) to appropriately adjudicate uncertainty.
The \methodNameAbbr protocol provides flexible, sample-efficient, and compute-efficient confidence estimators for 3 tasks including multiple choice question answering, tool-calling, and scientific document summarization across 6 models from 4 model families.
\methodNameAbbr outperforms strong baselines on \metricNameAbbr, while maintaining low calibration error.
Taken together, our work shows that equipping LLMs with the \methodNameAbbr protocol can improve calibration, utility, and trustworthiness in numerous settings.
\end{abstract}

\begin{document}
\frenchspacing
\maketitle

\section{Introduction}
\label{sec:introduction}

Users increasingly rely on large language models (LLMs) for many tasks such as information seeking, writing, and agentic tool-calling.
Many technological solutions integrate LLMs into their workflow at many stages, trusting and using model output for further computation.
As a result, automatically knowing whether to trust a model output is essential.

A confidence estimator can be an auxiliary model that estimates the probability that the output of another model (\eg an LLM) is correct.
\textit{Well-calibrated} confidence estimates can provide a signal of whether to trust a generation and what the associated risks could be. 
Practitioners often rely on a simple confidence estimator to evaluate whether to trust an LLM output: the probability the LLM itself assigns to the generation (\ie the product of token probabilities), even though this is known to be \textit{poorly calibrated} in both single- and multi-token generation settings~\citep{desai-durrett-2020-calibration, jiang-etal-2021-know, zhong-etal-2023-non, stengel-eskin-van-durme-2023-calibrated, subramani-etal-2025-mice}.

\begin{figure*}[t]
    \centering
    \includegraphics[width=\linewidth]{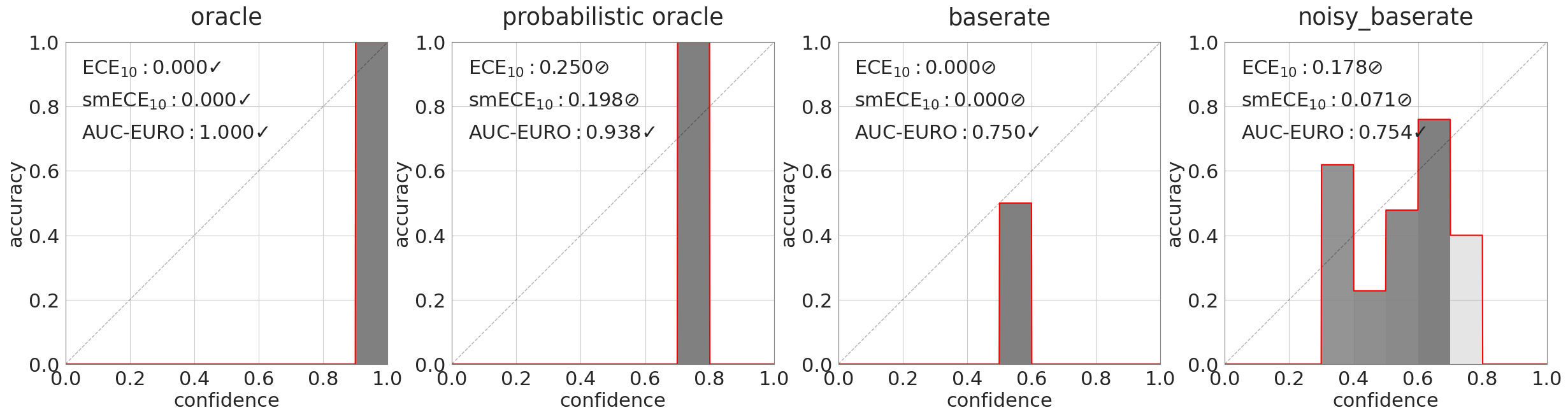}
    \caption{Here we show four different confidence estimators in order of decreasing quality and how the metrics ECE, smECE, and our metric \euroauc adjudicate them. Both ECE and smECE are errors so they should start at 0 and increase (\ie lower is better), while \euroauc is a utility so it should decrease (\ie higher is better). ECE and smECE fail, concluding that the two worst systems (baserate and noisy\_baserate) are \emph{much} better than the probabilistic oracle. \euroauc appropriately ranks them. Descriptions of each estimator are in~\cref{sec: calibration_alone_is_insufficient}.}
    \label{fig:euro_vs_ece}
\end{figure*}

Ideally, we want a metric to adjudicate how much to trust a model generation, while accounting for how risky the task is (\ie for a high risk task we require more certainty to trust an LLM output than a low risk task).
Calibration, often measured using expected calibration error (ECE;~\citealp{murphy1967verification, naeini2015obtaining}), solely measures how well predicted probabilities match observed outcomes.
This has \emph{two major drawbacks} for trust:

\textit{ECE cannot distinguish between an oracle and a perfectly uninformative estimator that just predicts the base rate for all outputs independent of correctness.}
Suppose we have a task that an LLM gets exactly 50\% accuracy on.
Both an oracle estimator (assigns a probability of 1 to correct and 0 to incorrect outputs) and a baserate estimator (assigns the base rate of 0.5 to all outputs) achieve an ECE of 0~(\cref{fig:euro_vs_ece}).

\textit{ECE is invariant to the associated risk of a task.}
Suppose there exist two tasks with different risks levels: high risk requiring 0.9 probability to trust and medium risk requiring only 0.5 to trust and a probabilistic oracle (\eg one that assigns 0.75 to correct and 0.25 to incorrect outputs).
This estimator gets an ECE of 0.25 for both tasks, even though it perfectly solves the medium-risk task, and fails on the high risk task (never trusting a single output).

Our contributions in this paper are:
\begin{enumerate}[nolistsep, noitemsep]
\item A new general and easily interpretable metric called \textbf{\metricName} (\metricNameAbbr), which balances calibration and decision-making utility.
\metricNameAbbr resolves both drawbacks of ECE, granting practitioners a more reliable trust estimator ~(\cref{sec:euro}).
\item An \textbf{activation-based confidence, utility, and trust estimation protocol} (\methodNameAbbr) to assess the confidence of language model outputs (\cref{sec:acute}).
Experiments on multiple-choice question answering (MMLU;~\citealp{Hendrycks2020MeasuringMM}), tool-calling (APIGen;~\citealp{liu2024apigen}), and scientific document summarization (SCITLDR;~\citealp{cachola-etal-2020-tldr}) show that applying the \methodNameAbbr protocol on 6 LLMs improves confidence estimation via \metricNameAbbr, while maintaining low calibration error.
\end{enumerate}

\section{\metricNameCaps (\metricNameAbbr)}
\label{sec:euro}

\subsection{Calibration, ECE, and Smooth ECE}
Calibration measures how well predicted probabilities match the observed outcomes for a task.
To measure how good a confidence estimator is for a task, expected calibration error (ECE) is often used~\citep{murphy1967verification, dawid1982well, naeini2015obtaining, naeini2016binary}.
To calculate ECE, we build a histogram where bins correspond to predicted confidence bands (\eg 3 bins equate to 0.0 - $\frac{1}{3}$, $\frac{1}{3}$ - $\frac{2}{3}$, $\frac{2}{3}$ - 1.0).
For every example $i$ within a bin, the accuracy of that example is compared to the estimator's predicted confidence: $\lvert acc_{i} - \hat{p_{i}}\rvert$.
All of these absolute differences are averaged to produce a final ECE score.
The number of bins can be challenging to set and different choices lead to different calibration scores.

\citet{blasiok2024smooth} introduce smooth ECE (smECE) a hyperparameter-free method to remove the reliance on the number of bins. Kernel width is determined automatically from data.
SmECE smooths out the histogram using Nadaraya-Watson kernel regression and measures calibration more precisely than ECE, but suffers from the same major drawbacks that ECE does. 
We introduce a metric that balances calibration with informativeness called \metricName (\metricNameAbbr) to assess the decision-making utility of a confidence estimator.
\metricNameAbbr has \emph{just one} parameter: the normalized net utility of correctly abstaining from trusting an incorrect generation (\normuca). 

\subsection{Calibration Alone is Insufficient}
\label{sec: calibration_alone_is_insufficient}
Suppose for some task, we have language model $M$ that outputs predictions $y_i, \ldots, y_n$ and has 50\% accuracy on these (\ie $\sum_i^n y_i = \frac{n}{2}$).
Consider the following estimators:
\begin{enumerate}[nolistsep]
    \item oracle (1 if $y_i$ is correct and 0 otherwise)
    \item probabilistic oracle (0.75 correct, 0.25 incorrect)
    \item baserate (0.5 regardless of what $y_i$ is)
    \item noisy baserate ($p\sim U(0.25, 0.75)$ regardless of $y_i$)
\end{enumerate}
The oracle is perfect and the probabilistic oracle is indistinguishable from the oracle as long as the probability threshold $\tau \in (0.25, 0.75]$ (\ie for $\hat{p}$ emit by the estimator, trust if $\hat{p} > \tau$).
Both base rate and noisy base rate estimators are uninformative for decision-making: they generate middling probabilities that are uncorrelated with correctness.

In~\cref{fig:euro_vs_ece}, we plot the reliability diagrams for these four estimators and their associated ECE, smECE, and our metric \euroauc, and find two concerning trends with both ECE and smECE.~\footnote{See~\cref{para:area_under_euro} for details on how to calculate \euroauc.}
First, they cannot distinguish between the oracle estimator and the perfectly uninformative base rate estimator (ECE=smECE=0).
Our metric, \euroauc, however, does distinguish between them (oracle gets a score of 1, while the baserate estimator gets a score of 0.75).
Second, the probabilistic oracle gets the worst performance score on ECE and smECE despite being a great confidence estimator.
\euroauc appropriately scores this as worse than the true oracle, but much better than either the baserate or noisy baserate estimators.

Both ECE metrics have a third drawback: they are invariant to the associated risk of a task.
Consider two tasks with different risk levels: a high risk task (\eg evaluating a trading agent that can buy and sell equities in a stock market) and a medium risk task (\eg scientific document summarization to use in a literature review for a course project).
For the high risk task, we must be very sure the output is correct before trusting the agent to execute a trade; lets suppose that the threshold $\tau$ to trust a generation is 0.9 (\ie $\hat{p} > 0.9$ to trust). 
For the medium risk task, lets assume that $\tau = 0.5$.
The probabilistic oracle estimator is good at abstaining for the high risk task, but it never trusts the output because all probabilities are less than 0.9.
An ideal metric should give the estimator a lot of credit for abstaining often, but with some penalty for never trusting.
For the medium risk task however, the probabilistic oracle estimator is perfect.
Since both calibration error metrics are invariant to risk level, the estimator gets the same scores regardless of task.
\metricNameAbbr, on the other hand, incorporates risk level; the probabilistic oracle achieves \metricNameAbbr= 0.9 and \metricNameAbbr= 1.0 for the high and medium risk tasks respectively.

\subsection{\metricNameAbbr}
For a given query $q$, a language model $M$ generates a candidate solution $\hat{y}$; a confidence estimator $C$ assigns a probability $\hat{p}$ to the example. There are four outcomes based on the Minimum Bayes Risk (MBR) decision: 
\begin{enumerate}[nolistsep]
    \item True Positives (tp) := $\hat{p} > \tau$ and $\hat{y}$ is correct
    \item False Positives (fp) := $\hat{p} > \tau$ and $\hat{y}$ is incorrect
    \item False Negatives (fn) := $\hat{p} \leq \tau$ and $\hat{y}$ is correct
    \item True Negatives (tn) := $\hat{p} \leq \tau$ and $\hat{y}$ is incorrect
\end{enumerate}

Each of these four outcomes has an associated reward or utility: \rtp, \rfp, \rfn, \rtn that can be set by the user based on the task at hand. Given the number of true positives, false positives, false negatives, and true negatives ($N_{tp}, N_{fp}, N_{fn}, N_{tn}$), the total utility $\mathbb{U} = R_{tp} * N_{tp} + R_{fp} * N_{fp} + R_{fn} * N_{fn} + R_{tn} * N_{tn}$. As a result, the MBR decision threshold for $\hat{p}$ is defined as:~\footnote{See the Appendix~\cref{sec:mbr_derivation} for a detailed derivation.}

\noindent
\begin{align}
\hat{p} > \tau & \eqdef \frac{R_{tn} - R_{fp}}{(R_{tp} - R_{fn}) + (R_{tn} - R_{fp})}
    \label{eqn:tau}
\end{align}

In our setting, the confidence estimator trusts the output generation if $\hat{p} > \tau$, and abstains if $\hat{p} \leq \tau$ because it is Bayes-optimal to do so.
Since a confidence estimator should get a greater reward for trusting when the generation is correct and abstaining when incorrect than the opposite, \rtp $>$ \rfn and \rtn $>$ \rfp.~\footnote{Often \rtp, \rtn $\in [0, \infty)$ and \rfn, \rfp $\in (-\infty, 0]$ because getting a classification decision correct generally yields a positive reward and getting it incorrect yields at most 0 reward. In a standard binary classification task \rtp $=$ \rtn $= 1$ and \rfp $=$ \rfn $= 0$ and thus $\tau = 0.5$.}
Setting four values that have large ranges can be challenging, unintuitive, and difficult to analyze.
To remedy this, we introduce two intuitive and tightly bounded quantities that help reparametrize this setup without loss of generality to have \emph{just one} degree of freedom.

\paragraph{Reparametrization:}
We are interested in comparing confidence estimators $C_1$ and $C_2$ on the outputs of the same language model $M$ on the same dataset $D$.
Crucially, $M$ generates $\hat{y}_1 \ldots \hat{y}_n$ and $C_1$ and $C_2$ \emph{both} estimate the confidences of these candidate generations.
\rtp, \rtn, \rfp, and \rfn are based on the dataset, not the LLM or confidence estimator.
When comparing $C_1$ and $C_2$ and the probabilities they emit, we only care about $\tau$ from~\cref{eqn:tau}.

\paragraph{Net Utility of Correctly Trusting (\uct):}
When the model generates a correct answer ($\hat{y}$ is correct), we want our confidence estimator to emit $\hat{p} > \tau$.
We assign a reward \rtp or \rfn based on whether $\hat{p} > \tau$ or not.
Recall that \rtp $>$ \rfn.
Our reformulation looks at their difference, the net utility of trusting a correct generation.

\paragraph{Net Utility of Correctly Abstaining (\uca):}
Similarly when $M$ generates an incorrect answer, we want the confidence estimator $C$ to output $\hat{p} \leq \tau$ and assign \rtn or \rfp accordingly.
We look again to their difference, the net utility of abstaining from trusting an incorrect generation. 
\[
    U_{ct} \eqdef R_{tp} - R_{fn}; U_{ca} \eqdef R_{tn} - R_{fp}
\]
\paragraph{Normalizing and Recomputing the MBR Threshold:}
To simplify further, we calculate a normalized version of both utilities \normuct, \normuca that are bounded between 0 and 1:
\begin{align}
\begin{bmatrix} u_{ct} \\ u_{ca} \end{bmatrix} &= \begin{bmatrix} U_{ct} \\ U_{ca} \end{bmatrix} \cdot \frac{1}{U_{ct} + U_{ca}}
\label{eqn:uct_uca}
\end{align}

Using \normuct and \normuca,~\cref{eqn:tau} can then be rewritten as:~\footnote{Given the normalization scheme we chose, \normuct = 1 - \normuca.}
\noindent
\begin{align}
    \hat{p} > \tau \eqdef \frac{U_{ca}}{U_{ct}+U_{ca}} = u_{ca}
    \label{eqn:tau_reparametrized}
\end{align}
We can now use $\tau$ as precisely the risk level associated with a task (\ie high risk means that an estimator gets a larger penalty for a false positive than reward for a true positive). 

\paragraph{Calculating the Relative Utility:}
To calculate the relative utility between two confidence estimators $C_1$ and $C_2$, remember that the rewards (\rtp, \rtn, \rfp, and \rfn) and language model generations ($\hat{y}_1, \ldots \hat{y}_n$) are fixed.
The only differences are the confusion matrices, \ie the counts of true positives, true negatives, false positives, and false negatives between the two systems based on~\cref{eqn:tau_reparametrized}.
The relative utility between two systems is the difference in total utility, $\mathbb{RU}_{C_1, C_2} = \mathbb{U}_{C_1} - \mathbb{U}_{C_2}$.
The fixed generations ensure that the number of correct and incorrect samples are preserved: $N_{tp, C_1} + N_{fn, C_1} = N_{tp, C_2} + N_{fn, C_2}$ and $N_{tn, C_1} + N_{fp, C_1} = N_{tn, C_2} + N_{fp, C_2}$.
Thus, simplifying and normalizing with respect to \normuca and \normuct yields:~\footnote{See Appendix~\cref{sec:ru_derivation} for all the \metricNameAbbr derivations.}
\begin{align}
    \mathbb{RU}_{C_1, C_2} = \frac{1}{U_{ct} + U_{ca}} \cdot
     \begin{bmatrix}
        u_{ct} \\ u_{ca}
    \end{bmatrix} \cdot \begin{bmatrix}
        (N_{tp, C_1} - N_{tp, C_2}) \\ (N_{tn, C_1} - N_{tn, C_2})
    \end{bmatrix}
    \label{eqn:relative_utility}
\end{align}

\paragraph{Renormalizing with respect to the Oracle and Anti-Oracle:}
We want to put relative utilities on a scale with respect to the Oracle policy ($O$), a system that only gets true positives and true negatives, and the Anti-Oracle policy ($AO$), a system that only gets false positives and false negatives.
Since the Oracle policy should get the highest possible utility and the Anti-Oracle policy get the lowest possible utility, we normalize the utility of an estimator accordingly:
$\metricNameAbbr_{C}(u_{ca}) \in [0,1]$, approaching 0 as $C$ gets closer to the $AO$ policy and approaching 1 as $C$ gets closer to the $O$ policy.
For a confidence estimator $C$, \metricNameAbbr is:
\begin{align}
    \metricNameAbbr_{C}(u_{ca}) = \frac{N_{tp, C} + u_{ca}\cdot(N_{tn, C} - N_{tp, C})}{N_{tp, O} + u_{ca}\cdot(N_{tn, O} - N_{tp, O})}
\end{align}

\paragraph{Area under the \metricNameAbbr curve}
\label{para:area_under_euro}
Since \metricNameAbbr is parametrized by \normuca, each value is a setting for the net utility of correctly abstaining from trusting an incorrect candidate.
We can compute the \metricNameAbbr for each of these, which traces a curve.
To compare confidence estimators across all possible settings, we summarize \metricNameAbbr by taking the area under this curve and term this \euroauc. Since \metricNameAbbr(\normuca) $\in [0,1]$, $\euroauc \in [0,1]$.~\footnote{AUC metrics are used widely in science (\eg ~\citealp{Wagner1977BioavailabilityAM, geifman2019biasreduceduncertaintyestimationdeep,subramani-etal-2025-simba}).}

\paragraph{Analyzing \metricNameAbbr across risk levels:}
In~\cref{eqn:tau_reparametrized}, we observe that $\tau \eqdef$\normuca, so we split \normuca into three risk bins, where low corresponds to \normuca$\in (0, \frac{1}{3})$, medium to \normuca$\in [\frac{1}{3}, \frac{2}{3})$, and high to \normuca$\in [\frac{2}{3}, 1)$.
For the highest risk tasks, we should be most weary of trusting a generation and get the most credit for correctly abstaining from trusting an incorrect generation.
For the lowest risk tasks, we should be the least weary of trusting a generation, but as a result, we get the least credit for correctly abstaining from trusting an incorrect generation.
The threshold $\tau$ and thus \normuca exactly determines how much confidence we require to trust a generation.
We measure a total of 4 \euroauc values: one corresponding to the entire spectra of \normuca values and one for each of the 3 risk-split settings mentioned above.~\footnote{See~\cref{sec:appendix_other_calibration_scores} for a discussion on how \metricNameAbbr relates to other calibration and decision-making utility tools.}

\section{\methodNameAbbr}
\label{sec:acute}

\begin{wrapfigure}{r}{0.53\textwidth}
    \centering
    \includegraphics[width=\linewidth]{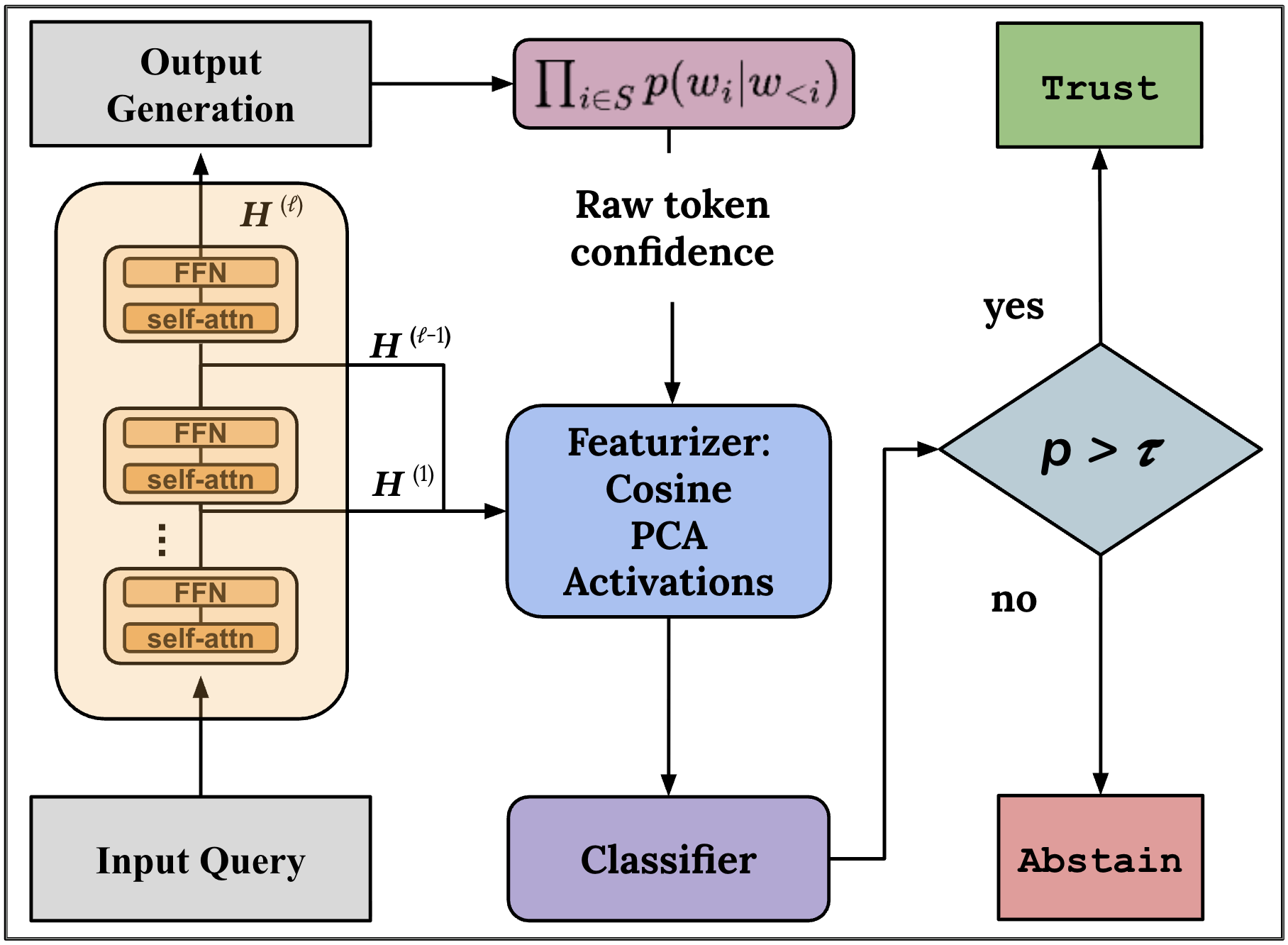}
    \caption{The \methodNameAbbr Protocol. We use input features from model internals to train confidence estimators to adjudicate whether to trust or abstain from trusting a model output. Details are in~\cref{sec:acute}.}
    \label{fig:acute_protocol}
\end{wrapfigure}

Activation-based confidence estimation leverages the activations of the language model while generating a candidate answer to learn a confidence estimator.
In particular, the activations are used as input features to a classifier that predicts whether the resulting generation is correct or not.
The probability that the classifier assigns to the "correct" class \emph{is} the new calibrated confidence.
The latent spaces of language models are steerable via steering vectors~\citep{subramani-etal-2022-extracting}, there exist circuits and individual directions in latent space that are controllable and task specific~\citep{li2023inference, conmy2023towards, syed-etal-2024-attribution, li2026circuitconsistency}, activations at different layers of models correspond to different linguistic phenomenon~\citep{tenney-etal-2019-bert, ethayarajh-2019-contextual, li2025model, acevedo2026differential}, model internals supercharge vocabulary expansion~\citep{mehta2026defragmenting}, and there exist interpretable latent structure~\citep{wurgaft2026manifoldsteeringrevealsshared}. 
Given these findings, we hypothesize that the activations contain decipherable signals for confidence estimation.

\subsection{Input features}
\paragraph{Mean-pooled Activations}
Since we suspect that the activation spaces contain decipherable information for confidence estimation, the raw activations themselves are the most natural place to start.
Consider a language model $M$ that processes an input query $q$ and produces a candidate generation $\hat{y} = w_1, \ldots, w_{T}$.
At each layer $j$, we extract the activations for the output sequence: $\hiddenStateAtLayer{j} = \begin{bmatrix}\hiddenTokenAtLayer{j}{1} & \ldots & \hiddenTokenAtLayer{j}{T}\end{bmatrix}$.
We collapse this activation into a single vector by mean pooling across the sequence such that for every layer $j$, we have a single activation vector per query: $\meanPoolHiddenStateAtLayer{j}$.
These become the input features to our confidence estimators.
We train separate confidence estimators at each layer, but present the performance of the best performing estimator at different layer depths (early, middle, and late) rather than at all layers for simplicity.

\paragraph{Layer-wise Cosine Similarities with the Final Layer}
Activations at a single layer are high-dimensional and can be limited in the information it could contain.
Layers tend to specialize; activations at earlier layers are better at probing word or subword-level phenomenon and those at layer layers are better at higher-level, sequence-level phenomenon~\citep{rogers-etal-2020-primer}. 
Since concatenating the representations from all layers is infeasible as the number of input features would be too large (\eg 200,000 input features for a model with a hidden size of 4000 and 50 layers), we choose an aggressive reduction. 
For a model with $\ell$ layers, we collect the mean pooled activations at each layer $\meanPoolHiddenStateAtLayer{1}, \ldots, \meanPoolHiddenStateAtLayer{\ell}$, 
and compute the cosine similarity between the pooled activation at each layer and that of the final layer:
\[%
    \cosineFeatures = \begin{bmatrix}\textsc{sim}(\meanPoolHiddenStateAtLayer{1}, \meanPoolHiddenStateAtLayer{\ell})~\ldots~\textsc{sim}(\meanPoolHiddenStateAtLayer{\ell-1}, \meanPoolHiddenStateAtLayer{\ell})\end{bmatrix} 
\]%
This aggregation method yields $\ell - 1$ total features, \ie $\cosineFeatures \in \mathbb{R}^{\ell-1}$.
We chose \textsc{cosine} as our similarity metric following previous work on aggregating embeddings to measure similarity, \eg BERTScore and CodeBERTScore~\citep{zhang2019bertscore, zhou-etal-2023-codebertscore}.
In practice, one could replace \textsc{cosine} with any other measure.
We use these $\ell - 1$ features to learn a confidence estimator.~\footnote{This estimator is a much more efficient version of the MICE estimators used to recalibrate tool-calling agents~\citep{subramani-etal-2025-mice}, see~\cref{sec:appendix_baselines} for more details.}

\paragraph{PCA Transformed Activations}
As mentioned earlier, the number of input features is still large, even when considering a single layer.
Aggregating across layers is infeasible without dimensionality reduction, but collapsing each layer down to just one feature like our cosine approach removes a lot of information.
We use Principal Components Analysis and compute the top $m$ components at each layer, leading to $m\ell$ features that we use to learn the estimator. We use 10 and 20 components in our experiments.

\subsection{\methodNameAbbr models}
We use these input features to train a simple classifier to predict whether a model generation is correct or not.
Random forests are strong correctness predictors for tool-calling and question-answering tasks~\citep{subramani-etal-2025-mice, liu2025llm}.~\footnote{In early experiments, we tested other simple classifiers like logistic regression and support vector machines, and found that random forests were slightly more performant.}
Traditionally, random forests require Platt scaling when used as confidence estimators as decision-trees favor extreme probabilities~\citep{platt1999probabilistic, zadrozny2001obtaining}, but~\citet{SHAKER2025114143} show that well-trained random forests do not need recalibration.
We experiment without recalibration in our experiments unless stated otherwise.
Hyperparameter details are in~\cref{sec:experiments}. 
\begin{table*}[t]
\footnotesize
\centering
\newcolumntype{d}{r@{.}l}
\newcommand\mc[1]{\multicolumn{1}{c}{#1}}
{\setlength{\tabcolsep}{3.5pt}
\begin{tabular}{l@{} c c c c c | c c c c c | c c c c c}
\toprule
& \multicolumn{5}{c}{\textbf{MMLU}} & \multicolumn{5}{c}{\textbf{APIGen}} & \multicolumn{5}{c}{\textbf{SCITLDR}}\\
\midrule
& ($\downarrow$) & \multicolumn{4}{c}{\textbf{\euroauc} ($\uparrow$)} & ($\downarrow$) & \multicolumn{4}{c} {\textbf{\euroauc} ($\uparrow$)} & ($\downarrow$) & \multicolumn{4}{c}{\textbf{\euroauc} ($\uparrow$)}\\
\cmidrule(r){2-6}
\cmidrule(r){7-11}
\cmidrule(r){12-16}
\textbf{estimator} & {\textbf{smECE}} & {low} & {med} & {high} & {all} & {\textbf{smECE}} & {low} & {med} & {high} & {all} & {\textbf{smECE}} & {low} & {med} & {high} & {all}\\
\midrule
{Raw Conf} & 0.17          & \underline{0.90} & 0.71 & 0.54 & 0.72 & 0.22 & 0.28 & 0.53 & 0.80 & 0.53 & 0.15 & 0.36 & \underline{0.71} & 	\textbf{0.92} & 0.66\\
{HRE}            & 0.11          & 0.87 & 0.72 & 0.71 & 0.77 & \textbf{0.02} & 0.82 & 0.70 & 0.82 & 0.78 & \textbf{0.08} & 0.67 & \underline{0.71} & \textbf{0.92} & \underline{0.77} \\
{\kernelReg}     & \textbf{0.07} & \underline{0.90} & 0.73 & 0.74 & 0.79 & \textbf{0.02} & 0.82 & 0.69 & 0.82 & 0.78 & \textbf{0.08} & 0.67 & \underline{0.71} & \textbf{0.92} & \underline{0.77} \\
\midrule
{\methodNameAbbr early act}  & \textbf{0.07} & \textbf{0.91} & 0.73 & 0.74 & 0.79 & 0.05 & \underline{0.90} & \underline{0.82} & \underline{0.87} & 0.86 & \textbf{0.08} & \underline{0.69} & \textbf{0.72} & \textbf{0.92} & \textbf{0.78}\\
{\methodNameAbbr mid act}    & \textbf{0.07} & \textbf{0.91} & \underline{0.76} & \underline{0.78} & \underline{0.82} & 0.06 & \underline{0.90} & \textbf{0.84} & \textbf{0.88} & \underline{0.87} & \textbf{0.08} & \textbf{0.70} & \textbf{0.72} & \textbf{0.92} & \textbf{0.78}\\
{\methodNameAbbr late act}   & \textbf{0.07}    & \textbf{0.91} & \textbf{0.77} & \textbf{0.80} & \textbf{0.83} & 0.07 & \underline{0.90} & \textbf{0.84} & \textbf{0.88} & \underline{0.87} & \textbf{0.08} & \underline{0.69} & \textbf{0.72} & \textbf{0.92} & \underline{0.77}\\
{\methodNameAbbr cosine}     & 0.09             & \underline{0.90} & 0.75 & \underline{0.78} & 0.81 & \underline{0.03} & 0.87 & 0.77 & 0.84 & 0.83 & \textbf{0.08} & 0.68 & \underline{0.71} & \textbf{0.92} & \underline{0.77} \\
{\methodNameAbbr pca10}      & \underline{0.08} & \textbf{0.91} & \underline{0.76} & \underline{0.78} & \underline{0.82} & 0.04 & \underline{0.90} & \underline{0.82} & \underline{0.87} & \underline{0.87} & \underline{0.09} & \textbf{0.70} & \textbf{0.72} & \textbf{0.92} & \textbf{0.78} \\
{\methodNameAbbr pca20}      & \underline{0.08} & \textbf{0.91} & \underline{0.76} & 0.77 & 0.81 & 0.06 & \textbf{0.91} & \textbf{0.84} & \textbf{0.88} & \textbf{0.88} & \underline{0.09} & \textbf{0.70} & \textbf{0.72} & \textbf{0.92} & \textbf{0.78} \\
 
\bottomrule
\end{tabular}
}
\caption{%
Results on the MMLU test set averaged across all 57 subtasks (left), on the APIGen test subset (middle), and on the SCITLDR dev set (right). All results for all tasks are averaged across the 6 LLMs we test. Lower smECE is better, while higher \euroauc is better. \textbf{Bold} indicates the best result and \underline{underline} indicates the second best result in each column.
}
\label{tab:results}
\end{table*}
\section{Experiments}
\label{sec:experiments}

We apply the \methodNameAbbr protocol on 6 models across 3 tasks and measure the degree to which calibration (via smECE) and utility (via \metricNameAbbr) improve over strong, efficient baselines.~\footnote{See~\cref{sec:appendix_baselines} for a more detailed discussion.}

\paragraph{Models}
In our experiments, we evaluate 6 different models that come from 4 model families: gemma-3-4b-it, gemma-3-12b-it~\citep{team2025gemma}, Qwen3-4B-Instruct-2507, Qwen3-14B~\citep{yang2025qwen3}, phi-4~\citep{abdin2024phi}, and SmolLM3-3B~\citep{bakouch2025smollm3}.
We apply the \methodNameAbbr protocol to each LLM separately and evaluate performance relative to strong baselines independently.

\paragraph{Datasets \& Tasks}
To measure how \methodNameAbbr generalizes, we choose three distinct tasks: language understanding (MMLU; 5-shot in-context learning), tool-calling (APIGen; zeroshot), and scientific document summarization (SCITLDR; zeroshot).
MMLU is a multiple choice English-language benchmark with 57 distinct subtasks ranging from elementary school mathematics and sciences to college and post-graduate level medicine and law~\citep{Hendrycks2020MeasuringMM}.
Crucially, these subtasks are all multiple choice, so LLMs generate a single token corresponding to the answer choice.
APIGen is a tool-calling dataset where the model is given a description and set of APIs to potentially call~\citep{liu2024apigen}.
We instruct the LLM to generate a tool call or set of tool calls with API names and parameters in a parsable format.
We experiment with the abstract only subtask of SCITLDR, where the model has to generate a short summary of approximately 25 words looking at just the abstract and title of the paper~\citep{cachola-etal-2020-tldr}.
Both APIGen and SCITLDR require generating multi-token output and are evaluated using approximate exact match (APIGen) and rouge-L score (SCITLDR).
Since we need binary labels for correctness, we use 0.3 as a threshold to stratify correct vs. incorrect generations for SCITLDR.~\footnote{See~\cref{sec:appendix_prompts} for system prompts,~\cref{sec: appendix_datasplits} for data splits, and~\cref{sec: scitldr_threshold_analysis} for an ablation over reasonable rouge-L threshold values.}
In all experiments, we use greedy decoding to control for stochasticity in the generation process and better connect raw confidences (joint probability of an output sequence under the LM) to the output generations themselves.~\footnote{\methodNameAbbr can easily be applied to model generations computed via other decoding algorithms (\eg nucleus sampling).}

\begin{figure*}[t]
\centering
    \includegraphics[width=\linewidth]{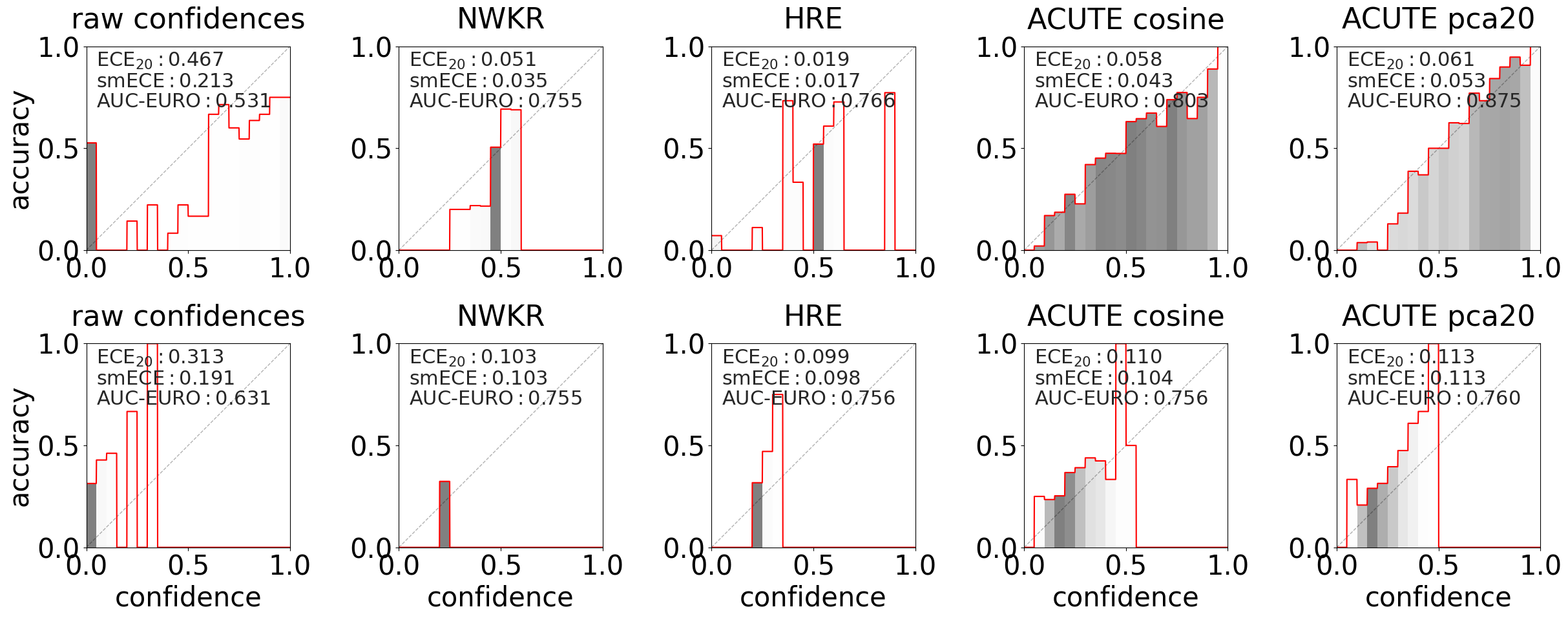}
    \caption{Here we show reliability diagrams for the gemma-3-12b-it model for APIGen (top row) and SCITLDR (bottom row) across 3 baseline estimators and two \methodNameAbbr estimators. Darker shading corresponds to higher density of examples in that confidence bin.}
    \label{fig: reliability_diagrams_gemma3}
\end{figure*}
\subsection{Baseline Confidence Estimators}

\paragraph{Raw Confidences:}
Raw confidences are the default baseline that our language models emit.
For an output sequence $y = w_1, \ldots, w_{T}$ of length $T$, the raw confidence is the joint probability that the language model $M$ assigns to $y$: $\hat{p} = \prod_{i=1}^{T} p_{M}(w_i | w_{<i})$.
Computing this does not require any supervised training data or training an auxiliary confidence estimator as these are emitted directly from the language model.
Raw confidence is used as an auxiliary feature for all of our \methodNameAbbr estimators because in early experiments, we saw a small performance gain when added.
See~\cref{sec: confidence_feature_ablation} for a more detailed discussion.

\paragraph{Histogram Regression Estimator (HRE):}
Our second baseline, HRE, is a standard recalibration method.
We construct a histogram of confidences based on a training set with equal sized bins.
To recalibrate a raw confidence estimate, we look up the bin associated with that estimate and return the average accuracy of all the samples in that bin. See~\citet{subramani-etal-2025-mice} for more details.

\paragraph{Nadaraya-Watson Kernel Regression (NWKR):}
For our third baseline, we use the Nadaraya-Watson kernel regression estimator~\citep{nadaraya1964estimating, watson1964smooth}, which is the method used to calculate smECE.
NWKR uses the raw confidence emitted by the language model and learns a reflected Gaussian kernel to map how over or under-confident a confidence estimate is.
Formally, to map from a confidence estimate emitted by the language model $p_M$ to a recalibrated confidence $\hat{p}$, we look up the NWKR estimate for $p_M$ and return that estimate: $\hat{p} = NWKR(p_M)$, like with HRE. 

\paragraph{Other Posthoc Calibration Baselines:}
We include three additional posthoc calibration baselines: Platt scaling~\citep{platt1999probabilistic}, isotonic regression~\citep{barlow1972statistical}, and beta calibration~\citep{pmlr-v54-kull17a}.
All of these methods recalibrate a single probability estimate by fitting a function on held-out calibration set.
Platt scaling fits a sigmoid, isotonic regression fits a step function with no defined functional form, and beta calibration fits a family of sigmoid functions via the beta cumulative distribution function.

\subsection{\methodNameAbbr Estimators}
We experiment with the different versions of the \methodNameAbbr estimators discussed in~\cref{sec:acute}: early activations, middle activations, late activations, cosine, pca10, and pca20.
As mentioned earlier, \methodNameAbbr early, medium, and late are just the highest performing classifier using mean-pooled activations as input features over the first, second, and final third of the layers (\ie for a 36 layer model, for \methodNameAbbr early, we look at the classifiers trained on each of the first 12 layers' activations and present the best performing one).
\methodNameAbbr cosine uses the layer-wise cosine similarities with the final layer as input features to learn a confidence estimator.
\methodNameAbbr pca10 and pca20 first PCA transform the activations at each layer to have 10 or 20 components and aggregate those across layers to produce input features.
For all \methodNameAbbr models, we train a random forest classifier on these different features to predict whether the input query is correct or not.
Our random forest uses 500 estimators, a max depth of 10, and does not restrict the number of features that can be used in each tree.
Note that \methodNameAbbr estimators are \emph{highly efficient}.
For Qwen3-30B-A3B-Instruct-2507 on APIGen, generation took an average of 15.75 seconds per example on 2 L40S GPUs. The amortized time for inference on a trained random forest (\methodNameAbbr pca20) for an example is 0.00007 seconds, so the wall-clock overhead is \emph{negligible} even if generation can be dramatically sped up.

\section{Results \& Analysis}
\label{sec:results}

\subsection{Main Results}
We present the results of evaluating the confidence estimators on MMLU, APIGen, and SCITLDR and evaluate both smECE and \euroauc in~\cref{tab:results}.
We report 4 \euroauc values: low risk (\normuca $\in [0, \frac{1}{3}))$, medium risk (med; \normuca $\in [\frac{1}{3}], \frac{2}{3}))$, high risk (\normuca $\in [\frac{2}{3}, 1))$, and all (\normuca $\in [0, 1))$.

\begin{figure*}[t]
    \centering
    \includegraphics[width=\linewidth]{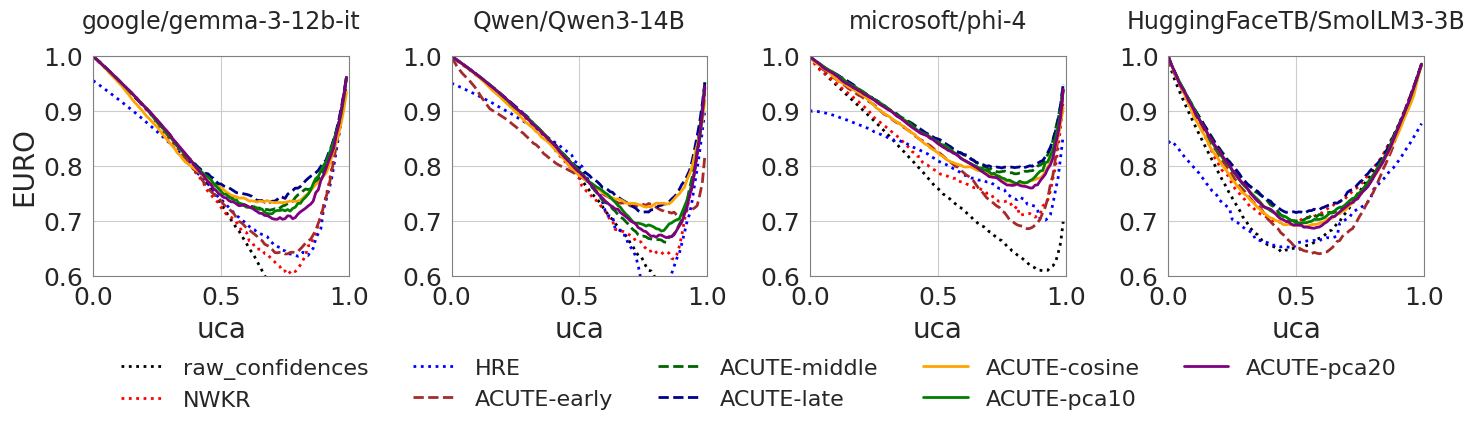}
    \caption{Results plotting \metricNameAbbr versus risk level (\normuca) for 4 of 6 LLMs (See~\cref{fig:full_mmlu_uplot} for the full plot). Results are averaged across the 57 subtasks of MMLU. We include 3 baselines (raw confidences, HRE, and NWKR) along with our 6 configurations of \methodNameAbbr. All policies perform similarly at the lowest risk levels, but at high risk settings, the \methodNameAbbr estimators perform best.}
    \label{fig:main_results}
\end{figure*}

Raw confidences have high calibration error and low \euroauc values across the board; it is the worst performing estimator across all tasks.
Inspecting the probabilities themselves (\cref{fig: reliability_diagrams_gemma3}), we observe rampant over-confidence in MMLU and under-confidence in APIGen and SCITLDR.
Both HRE and \kernelReg~improve upon the raw confidences by moving the probabilities into a much narrower range, whereby dramatically lowering smECE and increasing \euroauc.
Due to this, both baselines lead to low utility confidence estimates that make decision-making challenging, especially in high risk settings (see~\cref{fig:main_results} when \normuca $\in [0.4, 0.9]$).
In contrast, our \methodNameAbbr estimators have greater probability mass spread, while maintaining the low calibration error of HRE and \kernelReg, and having higher \euroauc scores at all risk levels on the tasks.

Our best \methodNameAbbr systems have statistically significant improvements ($p < 0.05$) on \euroauc over the best baselines on MMLU and APIGen across all models and on SCITLDR for 6 out of the 7 models (excluding phi-4). 
We use permutation tests with 1000 resamples to test the difference of means of \euroauc across our \methodNameAbbr systems and \kernelReg.
Additionally, the differences in smECE, where NWKR and HRE often outperform \methodNameAbbr, are not statistically significant ($p > 0.2$).

In~\cref{fig: reliability_diagrams_gemma3}, we show reliability diagrams for the gemma3-12b-it model on APIGen and SCITLDR.
Across tasks, HRE and NWKR bias for being a base rate estimator, whereas \methodNameAbbr systems have probability mass across the entire probability spectrum.
\methodNameAbbr offers recalibration with decision-making utility, whereas HRE and NWKR bias for calibration alone.
The best confidence estimators are \methodNameAbbr late and \methodNameAbbr pca20, having the highest \euroauc values across risk settings.
Given SCITLDR is a challenging task, NWKR, HRE, and all \methodNameAbbr methods perform comparably metric-wise, but baselines maintain degenerate behavior, nearly always predicting the base rate.
\methodNameAbbr always outperforms baselines on \euroauc, while maintaining smECE.
\subsection{Does \methodNameAbbr have strong sample complexity?}
\begin{figure}
\begin{minipage}[t]{0.55\textwidth}
    \centering
    \includegraphics[width=\linewidth]{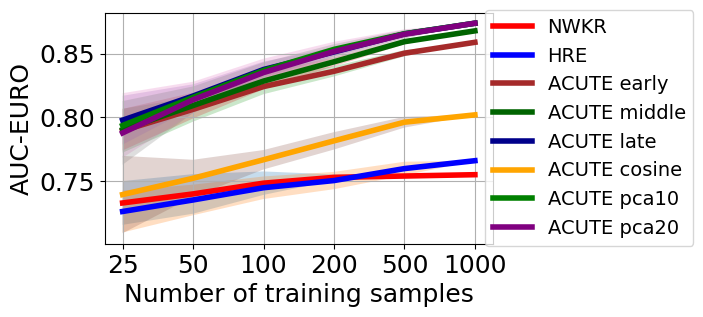}
    \caption{Sample complexity as the size of the training set varies of confidence estimators measured on the APIGen dataset on the gemma3-12b-it model. The error bands correspond to $\pm \sigma$.}
    \label{fig:sample_complex_gemma3-12b-it}
\end{minipage}
\hfill
\begin{minipage}[t]{0.43\textwidth}
    \centering
    \includegraphics[width=\linewidth]{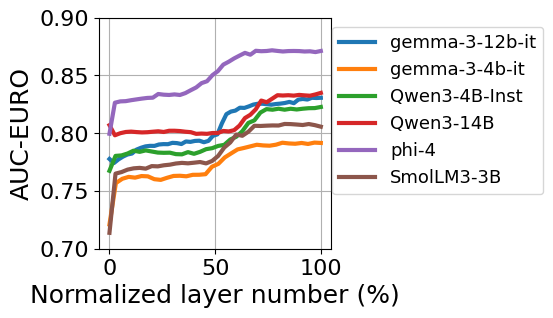}
    \caption{\euroauc performance of our mean-pooled activation \methodNameAbbr estimators on all 6 LLMs on MMLU across layers.}
    \label{fig:mmlu_layer}
\end{minipage}
\end{figure}

Here, we look at how sample efficient our methods are in comparison to the highly sample-efficient \kernelReg~baseline.~\footnote{~\citet{subramani-etal-2025-mice}, found that~\kernelReg~is highly sample-efficient because it uses just one feature.}
We look at effective training sizes $n_{i}$ = 25, 50, 100, 200, 500, 1000 on the APIGen dataset.
Recall the full training set has 1000 examples.
For each effective training set size $n_{i}$, we randomly sample $n_{i}$ examples without replacement from the overall training set 20 times and train different estimators for each of these 20 samples.
In~\cref{fig:sample_complex_gemma3-12b-it} we show the results of this for the gemma3-12b-it model with error bands corresponding to one standard deviation above and below the mean.
We find that all \methodNameAbbr methods outperform both baselines at all effective training set sizes, with the activation-based and pca-based estimators (\methodNameAbbr early, middle, late, \methodNameAbbr pca10, and \methodNameAbbr pca20) performing better with just 25 training examples than either baseline (\kernelReg, HRE) on all 1000 training examples.

\subsection{Which layers are best for \methodNameAbbr?}
One of our configurations for \methodNameAbbr is using mean-pooled activations at a specific layer as input features to our random forest classifier.
We dive deeper into this by looking at the performance of every layer across each of the 6 LLMs we tested on the MMLU benchmark, averaging results across all 57 tasks. 
In~\cref{fig:mmlu_layer}, we find that layerwise performance is very low at the first layer, indicating that the subword embeddings alone lack a strong confidence estimation signal.
Additionally, we find that there exist three phases that roughly correspond to the splits of early, middle, and late layers.
Early corresponds to $\sim$40\% of the layers, middle between 40 and 65\%, and late the remaining.
The early and late phases are flat, while the middle phase has a linearly increasing trend.
These trends are consistent across LLMs, suggesting that models tend to encode information useful for confidence estimation similarly across layers.

\subsection{Intrinsic Dimension Analysis}
To better understand how complex the representations extracted from our language models are, we perform an intrinsic dimension analysis using PCA.~\footnote{This is common practice to understand the effective dimension or intrinsic dimension of representations~\citep{li2018measuring, subramani2019can, li2025model}.}
On the APIGen dataset, we look at how many components (as a percentage of the total) are needed to explain 50\%, 70\%, 90\%, and 95\% of the variance in our test set.
Note that our test set has 1000 instances and our representation size for each of these models is larger than 1000.
Thus, trivially, 1000 components are sufficient to fully explain the variance in our data sample, so our maximum number of components is 1000.
\cref{fig:intrinsic_dim_apigen_gemma3-12b-it} shows the results of inspecting the activations of gemma3-12b-it model on APIGen and analyzing its intrinsic dimension.
We find that just a couple of components explain over 95\% of the variance for the first 60\% of the layers, but this increases at later layers perhaps providing better confidence estimation signal for our random forests.
These trends are consistent with all LLMs we tested.

\subsection{Does \methodNameAbbr generalize to larger MoE models?}
We test how our \methodNameAbbr methods scale to Qwen3-30B-A3B-Instruct-2507 on both APIGen and SCITLDR~\citep{yang2025qwen3}.
\cref{tab: apigen_scitldr_Qwen3-30B-A3B-Instruct-2507} shows good generalization; our earlier conclusions continue to hold for this model.
In fact, these results suggest that \methodNameAbbr may have a larger confidence estimation boost for larger mixture-of-experts models.

\begin{table*}[t]
\footnotesize
\centering
\newcolumntype{d}{r@{.}l}
\newcommand\mc[1]{\multicolumn{1}{c}{#1}}
{\setlength{\tabcolsep}{3.5pt}
\begin{tabular}{l@{} c c c c c | c c c c c }
\toprule
& \multicolumn{5}{c}{\textbf{APIGen}} & \multicolumn{5}{c}{\textbf{SCITLDR}}\\
\midrule
& & \multicolumn{4}{c} {\textbf{\euroauc} ($\uparrow$)} & & \multicolumn{4}{c}{\textbf{\euroauc} ($\uparrow$)}\\
\cmidrule(r){3-6}
\cmidrule(r){8-11}
\textbf{estimator} & {\textbf{smECE} ($\downarrow$)} & {low} & {med} & {high} & {all} & {\textbf{smECE} ($\downarrow$)} & {low} & {med} & {high} & {all} \\
\midrule
Raw Conf & 0.261 & 0.465 & 0.471 & 0.594 & 0.510 & 0.146 & 0.413 & 0.760 & \underline{0.939} & 0.701 \\
HRE & 0.028 & 0.941 & 0.799 & 0.718 & 0.820 & 0.088 & 0.637 & 0.763 & \textbf{0.940} & 0.779 \\
\kernelReg & 0.043 & 0.938 & 0.760 & 0.688 & 0.797 & \underline{0.083} & 0.641 & 0.763 & \textbf{0.940} & 0.780 \\
Platt & \underline{0.025} & 0.938 & 0.753 & 0.669 & 0.788 & \textbf{0.082} & 0.643 & 0.763 & \textbf{0.940} & 0.781 \\
Isotonic & \textbf{0.021} & 0.938 & 0.753 & 0.688 & 0.795 & \textbf{0.082} & 0.637 & 0.763 & \textbf{0.940} & 0.778 \\
Beta & 0.036 & 0.938 & 0.753 & 0.679 & 0.791 & \underline{0.083} & 0.641 & 0.761 & \textbf{0.940} & 0.779 \\
\midrule
\methodNameAbbr cosine & 0.031 & \underline{0.946} & 0.817 & 0.757 & 0.841 & 0.089 & 0.634 & 0.760 & \textbf{0.940} & 0.776 \\
\methodNameAbbr pca10 & \underline{0.025} & \underline{0.946} & 0.834 & 0.813 & 0.865 & 0.092 & 0.637 & 0.762 & \textbf{0.940} & 0.778 \\
\methodNameAbbr pca20 & 0.027 & \textbf{0.947} & 0.843 & 0.812 & 0.868 & 0.092 & \underline{0.653} & \textbf{0.765} & \textbf{0.940} & \textbf{0.785} \\
\methodNameAbbr early act & 0.047 & 0.942 & 0.825 & 0.784 & 0.851 & 0.089 & 0.652 & 0.763 & \textbf{0.940} & \underline{0.784} \\
\methodNameAbbr mid act & 0.057 & 0.943 & \underline{0.844} & \underline{0.821} & \underline{0.870} & 0.088 & \underline{0.653} & 0.763 & \textbf{0.940} & \underline{0.784} \\
\methodNameAbbr late act & 0.063 & 0.943 & \textbf{0.851} & \textbf{0.822} & \textbf{0.872} & 0.087 & \textbf{0.654} & \underline{0.764} & \textbf{0.940} & \underline{0.784} \\
\bottomrule
\end{tabular}
}
\caption{%
Results on the APIGen test subset (left), and on the SCITLDR dev set (right) on  Qwen3-30B-A3B-Instruct-2507. Lower smECE is better, while higher \euroauc is better. \textbf{Bold} indicates the best result and \underline{underline} indicates the second best result in each column.
}
\label{tab: apigen_scitldr_Qwen3-30B-A3B-Instruct-2507}
\end{table*}

\subsection{Can posthoc calibration further improve \methodNameAbbr?}
\begin{figure}
\begin{minipage}{0.39\textwidth}
    \centering
    \vspace{1em}
    \includegraphics[width=\linewidth]{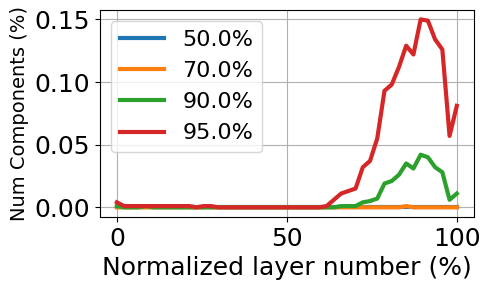}
    \caption{Intrinsic dimension of the activations for gemma-3-12b-it on the APIGen test set.}
    \label{fig:intrinsic_dim_apigen_gemma3-12b-it}
\end{minipage}
\hfill
\begin{minipage}{0.59\textwidth}
    \centering
    \includegraphics[width=\linewidth]{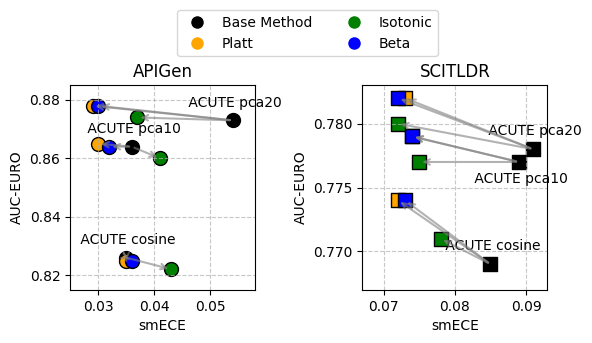}
    \caption{Effect of posthoc calibration with platt scaling, isotonic regression, and beta calibration on \methodNameAbbr systems on APIGEN (left) and SCITLDR (right). Posthoc calibration provides a reduction in calibration error without degrading \euroauc. Results here are averaged across the 7 LLMs evaluated in this paper.}%
    \label{fig:posthoc_calibrate_apigen_scitldr_average}
\end{minipage}
\end{figure}

In other words, are the gains with posthoc calibration (\eg Platt scaling, isotonic regression, and beta calibration) additive with \methodNameAbbr?
To test this, we posthoc calibrate the outputs of our \methodNameAbbr systems to reduce calibration error (smECE) on APIGen and SCITLDR.
~\cref{fig:posthoc_calibrate_apigen_scitldr_average} shows performance averaged across the 7 LLMs we tested.
We find that when applying posthoc calibration to the base systems (denoted in black), \euroauc generally increases and calibration error (smECE) decreases, especially for SCITLDR. For APIGen on \methodNameAbbr cosine, posthoc calibration has negligible impact with isotonic regression actually lowering performance.
Per model results are in~\cref{tab: apigen_scitldr_gemma3-12b-it,tab: apigen_scitldr_gemma3-4b-it,tab: apigen_scitldr_Qwen3-4B-Instruct-2507,tab: apigen_scitldr_Qwen3-14B,tab: apigen_scitldr_phi-4,tab: apigen_scitldr_SmolLM3-3B,tab: apigen_scitldr_Qwen3-30B-A3B-Instruct-2507_full,tab: apigen_scitldr_averaged}.

\section{Related Work}
\label{sec:related_work}

\paragraph{Calibration \& Confidence Estimation:}
Calibration has been studied in a variety of contexts often applied to binary classifiers.
Often these works consider only a single predicted confidence and include Platt scaling~\citep{platt1999probabilistic}, isotonic regression~\citep{barlow1972statistical}, beta calibration~\citep{pmlr-v54-kull17a}, and histogram-based adaptive binning methods~\citep{nobel1996histogram}.
Additionally different language technology systems have had their calibration assessed including large language models~\citep{jiang-etal-2021-know, kadavath2022language}, and machine translation systems~\citep{niculescu2005predicting}, while others leverage confidence estimates to ask for confirmation~\citep{stengel-eskin-van-durme-2023-mean}, and provide an estimate of trust to users~\citep{stengel2024lacie}.
Finally, as language model capabilities have improved, linguistic calibration, \ie asking the language model of its confidence, has become popular and has mixed utility~\citep{mielke-etal-2022-reducing, band2024linguistic}.
Prior work focus on single-token, classification-like settings, \methodNameAbbr, on the other hand, works with both single- and multi-token generation settings and offers a generalized activation-based calibration solution.

\paragraph{Model Internals:}
Many works have studied the internals of language models to identify how linguistic information is encoded throughout the layers of a model~\citep[\emph{inter alia}]{hewitt-manning-2019-structural, jawahar-etal-2019-bert,tenney-etal-2019-bert,niu-etal-2022-bert, he-etal-2024-decoding}. 
Language models can also be steered for exact generation~\citep{subramani2019can} and attribute-based steering~\citep{subramani-etal-2022-extracting, li2023inference}.
Others have found that there exist vectors that loosely correspond to functions or tasks~\citep{hendel-etal-2023-context, todd2023function}. 
Work combining mechanistic interpretability with calibration is rare:~\citet{beigi-etal-2024-internalinspector} use contrastive learning on model internals to improve trustworthiness,~\citet{subramani-etal-2025-mice} show that decoding from intermediate layers can improve calibration on tool-calling, and~\citet{liu2025llm} learn probes on activations to predict correctness in question-answering tasks.
\methodNameAbbr operationalizes the activation spaces to build compute- and sample-efficient confidence estimators to improve trustworthiness.
\section{Conclusion}
\label{sec:conclusion}
We identify major problems with calibration metrics when used for decision-making and propose a new general, interpretable metric called \textbf{\metricName} (\metricNameAbbr), which balances calibration and decision-making utility.
\metricNameAbbr can reliably score confidence estimators, resolving some drawbacks of ECE. 
Also, we introduce a light-weight, compute- and sample-efficient activation-based confidence, utility, and trust estimation protocol (\methodNameAbbr) to recalibrate LLM outputs. 
Experiments on MCQA (MMLU), tool-calling (APIGen), and scientific document summarization (SCITLDR) show that incorporating the \methodNameAbbr protocol on 6 LLMs improves confidence estimation (\metricNameAbbr), while maintaining low calibration error.
\section*{Impact Statement}
\label{sec:impact_statement}
In this paper, we present \metricNameAbbr and \methodNameAbbr.
\metricNameAbbr is a metric to measure how well calibrated and how useful a confidence estimator is for decision-making.
This gives users an alternative metric to measure how well their systems work across a variety of risk-levels.
\methodNameAbbr provides a light-weight, compute- and sample-efficient method to directly recalibrate LLM output probabilities.
Even though \methodNameAbbr requires access to model internals, the developers of a language model will always have access to its internals. In this competitive landscape, LLM developers are incentivized to improve user trust in their system, so incorporating \methodNameAbbr would be beneficial to that end.

Since better calibration improves decision-making for users, both of our contributions can help increase trust in language model outputs, either by helping to provide more reliable confidence estimates (through \methodNameAbbr) or by evaluating an already deployed confidence estimator (through \metricNameAbbr).

We believe that as LLMs become increasingly embedded in everyday technology, reliability and trust is paramount. 
We believe our work takes a small step to reiterate the need for better metrics and methods to appropriately adjudicate trust.
All that being said, our decision theoretic framework is not a replacement for careful data analysis and system design.

\section*{Acknowledgments}
We would like to thank Nivedita Suresh for feedback on~\cref{sec:euro}.
Additionally, we would like to thank the anonymous reviewers for numerous suggestions in the paper, including adding posthoc calibration methods as well as testing generalization to large MoE style models, which we did with Qwen3-30B-A3B-Instruct-2507.

\bibliographystyle{abbrvnat}
\nobibliography*
\bibliography{all_bibtextidy_abbr}

\newpage
\onecolumn
\appendix

\section{\metricNameAbbr derivations}
\label{sec:mbr_derivation}
Here we present some of the derivations from~\cref{sec:euro}.

\subsection{Bayes Optimality of the MBR Threshold}
Suppose a confidence estimator emits a probability $\hat{p}$ for an example.
From our vantage point, we have no other information than just this probability $\hat{p}$.
We expect that the probability that the example is correct is $\hat{p}$ and thus the probability that the example is incorrect is $1 - \hat{p}$.
For this task, lets assume we get the following rewards for true positives, true negatives, false positives, and false negatives: \rtp, \rtn, \rfp, and \rfn.
The reward associated with predicting the positive class (trusting the estimator) is: 
\begin{align}
    \hat{p} * R_{tp} + (1-\hat{p})* R_{fp}
\end{align}
The reward associated with predicting the negative class (abstaining from trusting the estimator) is:
\begin{align}
    (1-\hat{p}) * R_{tn} + (\hat{p})* R_{fn}
\end{align}

Given this, we should trust the estimator (predict the positive class) if and only if the reward for this is greater than the reward for abstaining (predicting the negative class):

\begin{align*}
    \hat{p} (R_{tp}) + (1-\hat{p}) R_{fp} &> (1-\hat{p}) R_{tn} + \hat{p} (R_{fn})\\
    \hat{p} (R_{tp} - R_{fn}) &> (1 - \hat{p}) (R_{tn} - R_{fp})\\
    \hat{p} (R_{tp} - R_{fn}) &> \hat{p} (R_{fp} - R_{tn}) + R_{tn} - R_{fp}\\
    \hat{p} (R_{tp} - R_{fn} - R_{fp} + R_{tn}) &> R_{tn} - R_{fp}\\
    \hat{p} &> \frac{R_{tn} - R_{fp}}{(R_{tp} - R_{fn} - R_{fp} + R_{tn})}\\
\end{align*}

We rearrange this to get the Minimum Bayes Risk threshold $\tau$ from~\cref{eqn:tau}:
\noindent
\begin{align}
\hat{p} > \tau & \eqdef \frac{R_{tn} - R_{fp}}{(R_{tp} - R_{fn}) + (R_{tn} - R_{fp})}
\end{align}.

\subsection{Recomputing the MBR Threshold}
Recall that $U_{ct} = R_{tp} - R_{fn}$ and $U_{ca} = R_{tn} - R_{fp}$.
Substituting the definitions of \uct, \uca, \normuct, and \normuca into~\cref{eqn:tau} yields:
\begin{align*}
\hat{p} > \tau & \eqdef \frac{R_{tn} - R_{fp}}{(R_{tp} - R_{fn}) + (R_{tn} - R_{fp})} = \frac{U_{ca}}{U_{ct} + U_{ca}} = u_{ca}\\
\end{align*}

\subsection{Calculating the Relative Utility}
\label{sec:ru_derivation}
Here we show the proof for~\cref{eqn:relative_utility}.
Recall that we are aiming to calculate the relative utility between two confidence estimators $C_1$ and $C_2$.
The rewards (\rtp, \rtn, \rfp, and \rfn) and language model generations ($\hat{y}_1, \ldots \hat{y}_n$) are fixed.
The confusion matrices are the only differences between the systems, so the only thing that varies are the counts of true positives, true negatives, false positives, and false negatives between the two systems based on~\cref{eqn:tau_reparametrized}.
We use the total reward calculation $\mathbb{U} = R_{tp} * N_{tp} + R_{tn} * N_{tn} + R_{fp} * N_{fp} + R_{fn} * N_{fn}$ to compute the relative utility between two systems.
This is precisely the difference in total utility: $\mathbb{RU}_{C_1, C_2} = \mathbb{U}_{C_1} - \mathbb{U}_{C_2}$.
Remember that our generations are fixed.
This ensures that the number of correct and incorrect samples are preserved: $N_{tp, C_1} + N_{fn, C_1} = N_{tp, C_2} + N_{fn, C_2}$ and $N_{tn, C_1} + N_{fp, C_1} = N_{tn, C_2} + N_{fp, C_2}$.
Rearranging this shows that $N_{tp, C_1} - N_{tp, C_2} = -N_{fn, C_1} + N_{fn, C_2}$ and $N_{tn, C_1} - N_{tn, C_2} = -N_{fp, C_1} + N_{fp, C_2}$.

\begin{align*}
    \mathbb{RU}_{C_1, C_2} &= \mathbb{U}_{C_1} - \mathbb{U}_{C_2} = \begin{bmatrix}
        N_{tp, C_1} - N_{tp, C_2} \\ N_{tn, C_1} - N_{tn, C_2} \\ N_{fp, C_1} - N_{fp, C_2} \\ N_{fn, C_1} - N_{fn, C_2} \\ 
    \end{bmatrix} \cdot \begin{bmatrix} R_{tp} \\ R_{tn} \\ R_{fp} \\ R_{fn} \end{bmatrix} \\
    \mathbb{RU}_{C_1, C_2}  &= \begin{bmatrix}
        N_{tp, C_1} - N_{tp, C_2} \\ N_{tn, C_1} - N_{tn, C_2} \\ -N_{tn, C_1} + N_{tn, C_2} \\ -N_{tp, C_1} + N_{tp, C_2} \\ 
    \end{bmatrix} \cdot \begin{bmatrix} R_{tp} \\ R_{tn} \\ R_{fp} \\ R_{fn} \end{bmatrix} \\
    \mathbb{RU}_{C_1, C_2} &= \begin{bmatrix}
        (R_{tp} - R_{fn}) \\ (R_{tn} - R_{fp})
    \end{bmatrix} \cdot \begin{bmatrix}
        (N_{tp, C_1} - N_{tp, C_2}) \\ (N_{tn, C_1} - N_{tn, C_2})
    \end{bmatrix}\\
    \mathbb{RU}_{C_1, C_2} &= \begin{bmatrix}
        U_{ct} \\ U_{ca}
    \end{bmatrix} \cdot \begin{bmatrix}
        (N_{tp, C_1} - N_{tp, C_2}) \\ (N_{tn, C_1} - N_{tn, C_2})
    \end{bmatrix}
\end{align*}
We can write this in terms of \normuct and \normuca, using~\cref{eqn:uct_uca}:
\begin{align}
    \mathbb{RU}_{C_1, C_2} = \frac{1}{U_{ct} + U_{ca}} \cdot
     \begin{bmatrix}
        u_{ct} \\ u_{ca}
    \end{bmatrix} \cdot \begin{bmatrix}
        (N_{tp, C_1} - N_{tp, C_2}) \\ (N_{tn, C_1} - N_{tn, C_2})
    \end{bmatrix} 
\end{align}

\subsection{Renormalizing with respect to the Oracle and Anti-Oracle Derivation}
\label{sec:euro_derivation}
Recall that we want our relative utilities to be scaled with respect to the Oracle ($O$) and Anti-Oracle ($AO$) policies, with $\metricNameAbbr_{O} = 1$ and $\metricNameAbbr_{AO} = 0$.
For a confidence estimator $C$, \metricNameAbbr is calculated as follows:
\begin{align}
    \metricNameAbbr_{C}(u_{ca}) 
    &= \frac{\mathbb{RU}_{C, AO}}{\mathbb{RU}_{O, AO}}
\end{align}
Notice that the $\frac{1}{U_{ct} + U_{ca}}$ is a constant in both relative utilities, so we can cancel those out.
\begin{align*}
    \metricNameAbbr_{C}(u_{ca}) &= 
    \begin{bmatrix}
        u_{ct} \\ u_{ca}
    \end{bmatrix} \cdot 
    \frac{\begin{bmatrix}
        (N_{tp, C} - N_{tp, AO}) \\ (N_{tn, C} - N_{tn, AO})
    \end{bmatrix}}{
    \begin{bmatrix}
        (N_{tp, O} - N_{tp, AO}) \\ (N_{tn, O} - N_{tn, AO})
    \end{bmatrix}} 
\end{align*}
The Anti-Oracle policy has no true positives or true negatives ($N_{tp, AO} = N_{tn, AO} = 0$). Also recall that \normuct = 1 - \normuca.
\begin{align*}
    \metricNameAbbr_{C}(u_{ca}) &= \frac{u_{ct} \cdot N_{tp, C} + u_{ca} \cdot N_{tn, C}}{u_{ct} \cdot N_{tp, O} + u_{ca} \cdot N_{tn, O}} = \frac{N_{tp, C} + u_{ca}\cdot(N_{tn, C} - N_{tp, C})}{N_{tp, O} + u_{ca}\cdot(N_{tn, O} - N_{tp, O})}\\
\end{align*}
\section{How \metricNameAbbr relates to other measures}
\label{sec:appendix_other_calibration_scores}

\metricNameAbbr is unique in that it combines calibration with decision-making utility together in one simple metric.
As mentioned earlier, calibration error metrics have major drawbacks that make decision-making challenging.
One of these drawbacks is the lack of incorporating the risk-level or the costs (and rewards) of true positives, false positives, true negatives, and false negatives into account.
An estimator that emits the same sets of probabilities for examples should be adjudicated differently for a high risk task versus a low risk one.
For a task with 0 risk, the right thing to do is to always trust your output regardless of whether it is correct.
As a result, any estimator that generates a nonzero probability would be scored equally.
Both calibration error metrics (ECE, smECE) lack this.
Brier score lacks this too, but has the guarantee that an objective function that minimizes it, will be Bayes-optimal.

Precision, recall, sensitivity, and specificity deal with the confusion matrix directly. 
The area under the precision recall curve (AUPRC) and the area under the receiver operating characteristic curve (AUROC) both involve summarizing the confusion matrix trading off these quantities.
In general, AUROC treats both classes as equal, so it is invariant to risk level.
AUPRC is not symmetric and is often used when there are imbalanced classes.
In a high (or low) risk scenario the rewards for the classes are imbalanced and thus AUPRC could be used.
However, both AUROC and AUPRC are reward-blind, so a MBR type of formulation needs to be used. \metricNameAbbr balances calibration and decision-making utility, while being able to appropriately score a confidence estimator across risk levels, taking into account calibration.

\section{Confidence Estimators can be Computationally Infeasible}
\label{sec:appendix_baselines}
Confidence estimators need to be light-weight and computationally-efficient to be useful in a variety of settings and actually be used in practice for language model generation.
Most methods for confidence estimation and model recalibration are expensive, nearly all applications ignore confidence estimation.
The ones that attempt recalibration rely on raw confidences or adhoc linguistic calibration on a small set of outputs.

Previous work on calibration recalibrates in an inefficient manner.
\citet{subramani-etal-2025-mice} propose MICE, which, to our knowledge, is the only model internals based confidence estimation work that leverages hidden states to recalibrate language model outputs in a single shot.
MICE requires logit lens decoding from each intermediate layer, which by itself is expensive, requiring additional matrix multiplies with the unembedding matrix for each layer.
Additionally, they compute the \textsc{bertscore} between the final output generation and each intermediate generation and requires using a large auxiliary language model (\texttt{DeBERTa-xlarge-mnli}) to extract these \textsc{bertscore} features before training their classifier.
This is computationally infeasible.
Even if the confidence estimator has been trained, each new inference query would require an additional forward passes of the \textsc{bertscore} model equal to the number of layers of the model along with the logit lens decoding overhead.
Assuming the underlying language model is the same size as \texttt{DeBERTa-xlarge-mnli}, each query takes 40-50x longer than the simple generation.

Our \methodNameAbbr cosine method approximates MICE.
We remove the reliance on the auxiliary model to compute \textsc{bertscore} by looking at the model's activations and comparing those with other layers.
This also removes the reliance on intermediate decoding, improving efficiency by at least 40x, and adding negligible overhead for confidence estimation.

\section{Connection to Selective Prediction}
From our understanding, a selective prediction system can decide to output a class label or abstain from it and is governed by a selective function. In our setting, our selective function is a combination of the confidence estimator and the minimum Bayes risk-based optimal policy. The selective classifier tries to minimize the selective risk at a given coverage value. In our setting, we define the risk level according to the normalized net utility of correctly abstaining (\normuca) based on the Bayes-optimal policy, and evaluate EURO according to that risk level. Given this, the risk-coverage curve defined in selective prediction literature and our \normuca vs. \metricNameAbbr curve are different.

\section{Prompts used for tasks}
\label{sec:appendix_prompts}
\subsection{MMLU}
\begin{lstlisting}[
    basicstyle=\small\ttfamily,
    breaklines=true,
    frame=single,
    caption={System Prompt for all models for MMLU}
]
"Answer the following multiple choice question by giving the most appropriate response. Answer with a single letter: either 'A', 'B', 'C', or 'D'. Even though the answer choices may have a period at the end, answer with just the letter. The following are multiple choice questions (with answers) about {CURR_SUBJECT} Question: {ICL EXAMPLE1} Answer: {ICL ANSWER1} ... Question: {ICL EXAMPLE 5} Answer: {ICL ANSWER5} Now answer the following multiple choice question with a single letter. Question: {QUESTION} Answer: "
\end{lstlisting}

\subsection{APIGen}
\begin{lstlisting}[
    basicstyle=\small\ttfamily,
    breaklines=true,
    frame=single,
    caption={System Prompt for all models for APIGen}
]
"You are an agent that specializes in function calling. You are given a list of tools, each with a name, description, and arguments. Each of the arguments may have a name, description, type, and a default value that the argument takes. You will be given a query. Please answer the query using the provided tools. Format your final answer as list of function calls in a json-like format without any linebreaks. Make sure the arugments for the function-calling are called arguments, not parameters. Here are the list of tools available: {CURRENT_EXAMPLES_AVAILABLE_TOOLS}. {INPUT_QUERY}"
\end{lstlisting}
\subsection{SCITLDR}
\begin{lstlisting}[
    basicstyle=\small\ttfamily,
    breaklines=true,
    frame=single,
    caption={System Prompt for all models for SCITLDR}
]
"You are a helpful assistant who is an expert in summarizing scientific papers. Given the title and abstract of the paper in the following format TITLE: <title> ABSTRACT: <abstract> SUMMARY: , write a short summary of the abstract, which will go after the SUMMARY: in the prompt. Summaries must be no more than one sentence, but this sentence can be longer than an average sentence. Most summaries are between 15 and 25 words. TITLE: {CURR_EXAMPLE_TITLE} ABSTRACT: {CURR_EXAMPLE_ABSTRACT} SUMMARY: "
\end{lstlisting}

\section{Dataset Information}
\label{sec: appendix_datasplits}
For MMLU, for each task we split into training and evaluation sets. If the task had less than 200 examples, we used 50 for training, less than 400 examples, 100 for training, and otherwise 200 for training. All other examples were put in the evaluation set. For APIGen we randomly assigned 1000 examples for training and 1000 for evaluation. For SCITLDR, we use the training set (1992 samples) for training and the development set (618 samples) for evaluation. When including the post-hoc calibration, we split each training set into 80\% training and 20\% calibration.

\section{Confidence Feature Ablation}
\label{sec: confidence_feature_ablation}

Here we show results from early experiments when adding raw confidence as a feature to our \methodNameAbbr systems. 
These are averaged across 6 LLMs we test (all models not including Qwen3-30B-A3B-Instruct-2507 and result in a negligibly positive differences.

\begin{table}[h]
\centering
\begin{tabular}{cccc}
\toprule
\textbf{Estimator} & \textbf{Dataset} & \textbf{$\Delta$smECE ($\downarrow$)} & {$\Delta$\euroauc ($\uparrow$)} \\
\midrule
\methodNameAbbr pca10 & MMLU (all) & 0.0   & 0.002 \\
\methodNameAbbr pca20 & MMLU (all) & 0.0   & 0.001 \\
\methodNameAbbr pca10 & APIGen     & 0.0   & 0.0 \\
\methodNameAbbr pca20 & APIGen     & 0.002 & 0.0 \\
\methodNameAbbr pca10 & SCITLDR    & 0     & 0.0 \\
\methodNameAbbr pca20 & SCITLDR    & 0.001 & 0.0 \\
\bottomrule
\end{tabular}
\caption{Here we show the difference in smECE and $\euroauc$ when adding the confidence feature for \methodNameAbbr pca10 and \methodNameAbbr pca20 across all three datasets. Deltas are very small indicating that the feature is not vital to include.}
\label{tab: confidence_feature_ablation}
\end{table}

\section{Noise Ablation}
To test whether any multi-variate vectors help with confidence estimation, we construct a noise matrix by sampling from a multi-variate gaussian parametrized by each feature’s mean and variance.
We interpolate between the original features and the noise matrix on both training and test sets and evaluate \methodNameAbbr pca20 via smECE and \euroauc averaged across all LLMs on APIGen.
N\% noise corresponds to $N*X_{noise} + (1-N)*X_{features}$
We include the base rate estimator (one that just predicts training set accuracy for all inputs). 
Results hint that the activations contain relevant signals for confidence estimation and that performance degrades to the base rate estimator with more noise.

\begin{table}[h]
\centering
\begin{tabular}{ccc}
\toprule
\textbf{Noise Level} & \textbf{smECE ($\downarrow$)} & {\euroauc ($\uparrow$)} \\
\midrule
0\%                 &  0.05   & \textbf{0.873} \\
25\%                &  0.06   & \underline{0.867} \\
50\%                &  0.06   & 0.850 \\
75\%                &  0.05   & 0.814 \\
100\%               &  0.05   & 0.778 \\
Base Rate Estimator &  0.001  & 0.777 \\
\bottomrule
\end{tabular}
\caption{Here we show the smECE and $\euroauc$ when adding different magnitudes of Gaussian noise to the input data for the \methodNameAbbr pca20 estimator on APIGen. Intuitively, increased noise degrades performance on decision-making utility.} 
\label{tab: gaussian_noise_ablation}
\end{table}

\section{MMLU results across LLMs}
\begin{figure*}[h]
    \centering
    \includegraphics[width=\linewidth]{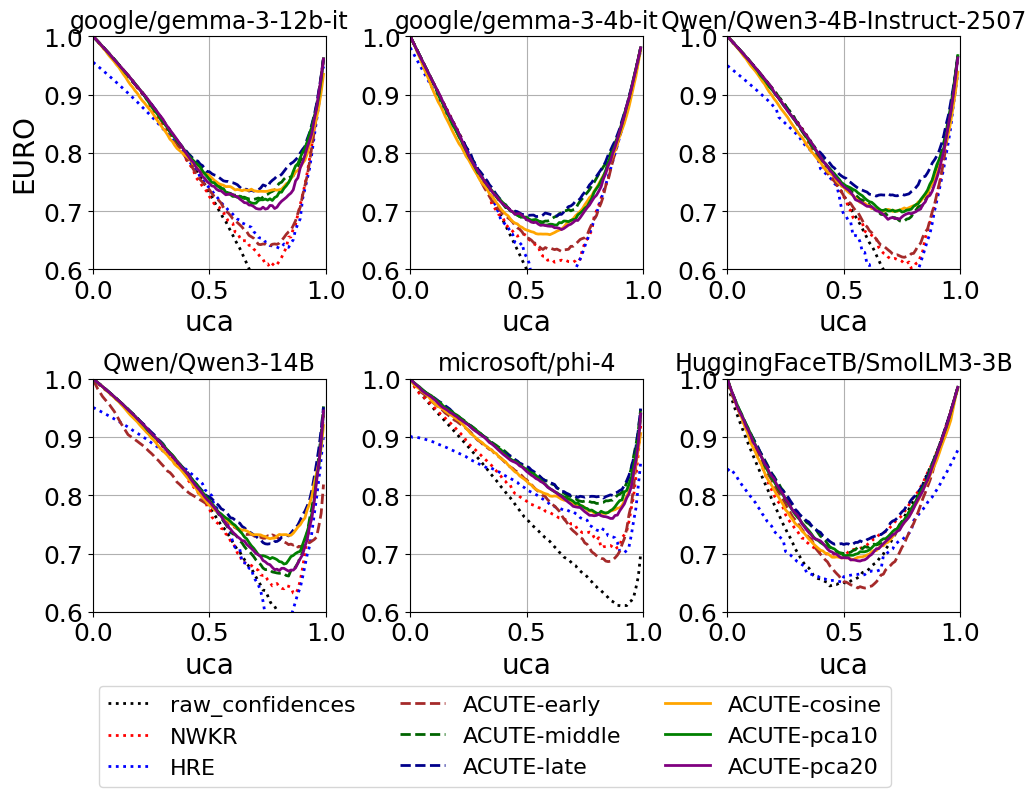}
    \caption{Results plotting \metricNameAbbr versus risk level (\normuca) for 6 LLMs. Performances are averaged across the 57 subtasks of MMLU. We include 3 baselines (raw confidences, HRE, and NWKR) along with our 6 configurations of \methodNameAbbr. All policies perform similarly at the lowest risk levels, but at high risk settings, the \methodNameAbbr estimators perform best.}
    \label{fig:full_mmlu_uplot}
\end{figure*}

In~\cref{fig:full_mmlu_uplot}, we show the full plot of results across all 6 LLMs. We find that trends are very consistent and show that \methodNameAbbr systems perform best at high risk levels and comparably to baselines at lower risk levels.

\section{SCITLDR Threshold Analysis}
\label{sec: scitldr_threshold_analysis}

In~\cref{tab:scitldr_threshold_results_gemma3-12b-it,tab:scitldr_threshold_results_gemma3-4b-it,tab:scitldr_threshold_results_Qwen3-4B-Instruct-2507,tab:scitldr_threshold_results_Qwen3-14B,tab:scitldr_threshold_results_phi-4,tab:scitldr_threshold_results_SmolLM3-3B,tab:scitldr_threshold_results_Qwen3-30B-A3B-Instruct-2507}, we show the per model results for varying the Rouge-L threshold to decide whether an example is correct or not across the LLMs we tested. Note the mean accuracies (correct vs. incorrect on the language model output) across  these LLMs for different rouge-L thresholds are approximately 60\% for 0.2, 22\% for 0.3, and 6\% for 0.4 with low variances across models.

\begin{table*}[t]
\footnotesize
\centering
\newcolumntype{d}{r@{.}l}
\newcommand\mc[1]{\multicolumn{1}{c}{#1}}
{\setlength{\tabcolsep}{3.5pt}
\begin{tabular}{l@{} c c c c c | c c c c c | c c c c c}
\toprule
& \multicolumn{5}{c}{\textbf{Threshold = 0.2}} & \multicolumn{5}{c}{\textbf{Threshold = 0.3}} & \multicolumn{5}{c}{\textbf{Threshold = 0.4}}\\
\midrule
& ($\downarrow$) & \multicolumn{4}{c}{\textbf{\euroauc} ($\uparrow$)} & ($\downarrow$) & \multicolumn{4}{c} {\textbf{\euroauc} ($\uparrow$)} & ($\downarrow$) & \multicolumn{4}{c}{\textbf{\euroauc} ($\uparrow$)}\\
\cmidrule(r){2-6}
\cmidrule(r){7-11}
\cmidrule(r){12-16}
\textbf{estimator} & {\textbf{smECE}} & {low} & {med} & {high} & {all} & {\textbf{smECE}} & {low} & {med} & {high} & {all} & {\textbf{smECE}} & {low} & {med} & {high} & {all}\\
\midrule
Raw Conf & 0.39 & 0.10 & 0.22 & 0.60 & 0.30 & 0.19 & 0.32 & 0.67 & 0.91 & 0.63 & 0.04 & 0.67 & 0.92 & 0.98 & 0.86 \\ 
HRE & 0.20 & 0.94 & 0.69 & 0.60 & 0.75 & 0.10 & 0.69 & 0.67 & 0.91 & 0.76 & 0.02 & 0.74 & 0.92 & 0.98 & 0.88 \\  
\kernelReg & 0.20 & 0.95 & 0.69 & 0.60 & 0.75 & 0.10 & 0.69 & 0.67 & 0.91 & 0.75 & 0.02 & 0.74 & 0.92 & 0.98 & 0.88 \\ 
Platt & 0.20 & 0.95 & 0.70 & 0.60 & 0.75 & 0.10 & 0.68 & 0.67 & 0.91 & 0.75 & 0.02 & 0.73 & 0.92 & 0.98 & 0.88 \\ 
Isotonic & 0.20 & 0.95 & 0.70 & 0.60 & 0.75 & 0.10 & 0.69 & 0.67 & 0.91 & 0.76 & 0.02 & 0.74 & 0.92 & 0.98 & 0.88 \\ 
Beta & 0.20 & 0.95 & 0.69 & 0.60 & 0.75 & 0.10 & 0.69 & 0.67 & 0.91 & 0.75 & 0.02 & 0.73 & 0.92 & 0.98 & 0.88 \\ 
\midrule
\methodNameAbbr cosine & 0.20 & 0.95 & 0.68 & 0.61 & 0.75 & 0.10 & 0.69 & 0.67 & 0.91 & 0.75 & 0.03 & 0.72 & 0.92 & 0.98 & 0.87 \\  
\methodNameAbbr pca10 & 0.20 & 0.95 & 0.68 & 0.61 & 0.75 & 0.11 & 0.70 & 0.68 & 0.91 & 0.76 & 0.03 & 0.72 & 0.92 & 0.98 & 0.87 \\  
\methodNameAbbr pca20 & 0.20 & 0.95 & 0.70 & 0.61 & 0.75 & 0.11 & 0.70 & 0.67 & 0.91 & 0.76 & 0.03 & 0.72 & 0.92 & 0.98 & 0.87 \\  
\methodNameAbbr early act & 0.20 & 0.95 & 0.71 & 0.60 & 0.75 & 0.11 & 0.70 & 0.67 & 0.91 & 0.76 & 0.02 & 0.74 & 0.92 & 0.98 & 0.88 \\  
\methodNameAbbr mid act & 0.20 & 0.95 & 0.71 & 0.60 & 0.76 & 0.10 & 0.70 & 0.67 & 0.91 & 0.76 & 0.02 & 0.74 & 0.92 & 0.98 & 0.88 \\  
\methodNameAbbr late act & 0.20 & 0.95 & 0.71 & 0.60 & 0.76 & 0.10 & 0.69 & 0.67 & 0.91 & 0.76 & 0.02 & 0.73 & 0.92 & 0.98 & 0.88 \\ \bottomrule
\end{tabular}
}
\caption{%
Results on the SCITLDR dev set evaluated at different Rouge-L thresholds to test threshold sensitivity on gemma3-12b-it. Lower smECE is better, while higher \euroauc is better. 
}
\label{tab:scitldr_threshold_results_gemma3-12b-it}
\end{table*}

\begin{table*}[t]
\footnotesize
\centering
\newcolumntype{d}{r@{.}l}
\newcommand\mc[1]{\multicolumn{1}{c}{#1}}
{\setlength{\tabcolsep}{3.5pt}
\begin{tabular}{l@{} c c c c c | c c c c c | c c c c c}
\toprule
& \multicolumn{5}{c}{\textbf{Threshold = 0.2}} & \multicolumn{5}{c}{\textbf{Threshold = 0.3}} & \multicolumn{5}{c}{\textbf{Threshold = 0.4}}\\
\midrule
& ($\downarrow$) & \multicolumn{4}{c}{\textbf{\euroauc} ($\uparrow$)} & ($\downarrow$) & \multicolumn{4}{c} {\textbf{\euroauc} ($\uparrow$)} & ($\downarrow$) & \multicolumn{4}{c}{\textbf{\euroauc} ($\uparrow$)}\\
\cmidrule(r){2-6}
\cmidrule(r){7-11}
\cmidrule(r){12-16}
\textbf{estimator} & {\textbf{smECE}} & {low} & {med} & {high} & {all} & {\textbf{smECE}} & {low} & {med} & {high} & {all} & {\textbf{smECE}} & {low} & {med} & {high} & {all}\\
\midrule
Raw Conf & 0.38 & 0.11 & 0.26 & 0.64 & 0.33 & 0.17 & 0.36 & 0.71 & 0.92 & 0.66 & 0.03 & 0.67 & 0.93 & 0.98 & 0.86 \\ 
HRE & 0.20 & 0.94 & 0.63 & 0.64 & 0.74 & 0.08 & 0.67 & 0.71 & 0.92 & 0.76 & 0.01 & 0.74 & 0.93 & 0.98 & 0.88 \\  
\kernelReg & 0.19 & 0.94 & 0.65 & 0.64 & 0.74 & 0.08 & 0.67 & 0.71 & 0.92 & 0.77 & 0.01 & 0.74 & 0.93 & 0.98 & 0.88 \\ 
Platt & 0.19 & 0.94 & 0.64 & 0.64 & 0.74 & 0.07 & 0.67 & 0.71 & 0.92 & 0.77 & 0.01 & 0.74 & 0.93 & 0.98 & 0.88 \\ 
Isotonic & 0.20 & 0.94 & 0.63 & 0.64 & 0.74 & 0.08 & 0.68 & 0.71 & 0.92 & 0.77 & 0.01 & 0.74 & 0.93 & 0.98 & 0.88 \\ 
Beta & 0.20 & 0.94 & 0.64 & 0.64 & 0.74 & 0.07 & 0.67 & 0.71 & 0.92 & 0.77 & 0.01 & 0.74 & 0.93 & 0.98 & 0.88 \\  
\midrule
\methodNameAbbr cosine & 0.19 & 0.93 & 0.64 & 0.64 & 0.74 & 0.08 & 0.65 & 0.71 & 0.92 & 0.76 & 0.03 & 0.72 & 0.92 & 0.98 & 0.88 \\  
\methodNameAbbr pca10 & 0.20 & 0.93 & 0.64 & 0.65 & 0.74 & 0.09 & 0.68 & 0.72 & 0.92 & 0.77 & 0.03 & 0.73 & 0.93 & 0.98 & 0.88 \\
\methodNameAbbr pca20 & 0.20 & 0.94 & 0.64 & 0.65 & 0.74 & 0.09 & 0.69 & 0.71 & 0.92 & 0.77 & 0.02 & 0.74 & 0.93 & 0.98 & 0.88 \\
\methodNameAbbr early act & 0.19 & 0.94 & 0.67 & 0.64 & 0.75 & 0.08 & 0.70 & 0.71 & 0.92 & 0.78 & 0.02 & 0.75 & 0.93 & 0.98 & 0.88 \\  
\methodNameAbbr mid act & 0.19 & 0.94 & 0.67 & 0.64 & 0.75 & 0.08 & 0.69 & 0.71 & 0.92 & 0.77 & 0.02 & 0.74 & 0.93 & 0.98 & 0.88 \\  
\methodNameAbbr late act & 0.19 & 0.94 & 0.66 & 0.64 & 0.75 & 0.08 & 0.69 & 0.71 & 0.92 & 0.77 & 0.02 & 0.74 & 0.93 & 0.98 & 0.88 \\ 
\bottomrule
\end{tabular}
}
\caption{%
Results on the SCITLDR dev set evaluated at different Rouge-L thresholds to test threshold sensitivity on gemma3-4b-it. Lower smECE is better, while higher \euroauc is better. 
}
\label{tab:scitldr_threshold_results_gemma3-4b-it}
\end{table*}

\begin{table*}[t]
\footnotesize
\centering
\newcolumntype{d}{r@{.}l}
\newcommand\mc[1]{\multicolumn{1}{c}{#1}}
{\setlength{\tabcolsep}{3.5pt}
\begin{tabular}{l@{} c c c c c | c c c c c | c c c c c}
\toprule
& \multicolumn{5}{c}{\textbf{Threshold = 0.2}} & \multicolumn{5}{c}{\textbf{Threshold = 0.3}} & \multicolumn{5}{c}{\textbf{Threshold = 0.4}}\\
\midrule
& ($\downarrow$) & \multicolumn{4}{c}{\textbf{\euroauc} ($\uparrow$)} & ($\downarrow$) & \multicolumn{4}{c} {\textbf{\euroauc} ($\uparrow$)} & ($\downarrow$) & \multicolumn{4}{c}{\textbf{\euroauc} ($\uparrow$)}\\
\cmidrule(r){2-6}
\cmidrule(r){7-11}
\cmidrule(r){12-16}
\textbf{estimator} & {\textbf{smECE}} & {low} & {med} & {high} & {all} & {\textbf{smECE}} & {low} & {med} & {high} & {all} & {\textbf{smECE}} & {low} & {med} & {high} & {all}\\
\midrule
Raw Conf & 0.37 & 0.16 & 0.27 & 0.65 & 0.36 & 0.16 & 0.38 & 0.71 & 0.92 & 0.67 & 0.03 & 0.74 & 0.94 & 0.99 & 0.89 \\
HRE & 0.18 & 0.93 & 0.63 & 0.66 & 0.74 & 0.07 & 0.66 & 0.72 & 0.92 & 0.77 & 0.01 & 0.77 & 0.95 & 0.99 & 0.90 \\  
\kernelReg & 0.18 & 0.93 & 0.64 & 0.65 & 0.74 & 0.08 & 0.67 & 0.71 & 0.92 & 0.77 & 0.01 & 0.78 & 0.95 & 0.99 & 0.90 \\
Platt & 0.18 & 0.93 & 0.64 & 0.65 & 0.74 & 0.07 & 0.67 & 0.71 & 0.92 & 0.77 & 0.01 & 0.78 & 0.95 & 0.99 & 0.90 \\
Isotonic & 0.18 & 0.93 & 0.65 & 0.65 & 0.75 & 0.07 & 0.68 & 0.71 & 0.92 & 0.77 & 0.01 & 0.78 & 0.95 & 0.99 & 0.90 \\
Beta & 0.18 & 0.93 & 0.65 & 0.65 & 0.75 & 0.07 & 0.67 & 0.71 & 0.92 & 0.77 & 0.01 & 0.78 & 0.95 & 0.99 & 0.90 \\
\midrule
\methodNameAbbr cosine & 0.18 & 0.93 & 0.64 & 0.66 & 0.74 & 0.08 & 0.68 & 0.71 & 0.92 & 0.77 & 0.03 & 0.76 & 0.95 & 0.99 & 0.90 \\  
\methodNameAbbr pca10 & 0.18 & 0.93 & 0.65 & 0.66 & 0.75 & 0.08 & 0.70 & 0.71 & 0.92 & 0.78 & 0.02 & 0.78 & 0.95 & 0.99 & 0.90 \\  
\methodNameAbbr pca20 & 0.18 & 0.93 & 0.67 & 0.66 & 0.76 & 0.08 & 0.70 & 0.72 & 0.92 & 0.78 & 0.02 & 0.78 & 0.95 & 0.99 & 0.90 \\  
\methodNameAbbr early act & 0.17 & 0.93 & 0.67 & 0.66 & 0.76 & 0.08 & 0.69 & 0.71 & 0.92 & 0.77 & 0.01 & 0.79 & 0.95 & 0.99 & 0.91 \\  
\methodNameAbbr mid act & 0.17 & 0.93 & 0.67 & 0.66 & 0.76 & 0.08 & 0.69 & 0.71 & 0.92 & 0.77 & 0.01 & 0.79 & 0.95 & 0.99 & 0.91 \\  
\methodNameAbbr late act & 0.18 & 0.93 & 0.67 & 0.66 & 0.75 & 0.08 & 0.68 & 0.71 & 0.92 & 0.77 & 0.01 & 0.78 & 0.95 & 0.99 & 0.91 \\ 
\bottomrule
\end{tabular}
}
\caption{%
Results on the SCITLDR dev set evaluated at different Rouge-L thresholds to test threshold sensitivity on Qwen3-4B-Instruct-2507. Lower smECE is better, while higher \euroauc is better. 
}
\label{tab:scitldr_threshold_results_Qwen3-4B-Instruct-2507}
\end{table*}

\begin{table*}[t]
\footnotesize
\centering
\newcolumntype{d}{r@{.}l}
\newcommand\mc[1]{\multicolumn{1}{c}{#1}}
{\setlength{\tabcolsep}{3.5pt}
\begin{tabular}{l@{} c c c c c | c c c c c | c c c c c}
\toprule
& \multicolumn{5}{c}{\textbf{Threshold = 0.2}} & \multicolumn{5}{c}{\textbf{Threshold = 0.3}} & \multicolumn{5}{c}{\textbf{Threshold = 0.4}}\\
\midrule
& ($\downarrow$) & \multicolumn{4}{c}{\textbf{\euroauc} ($\uparrow$)} & ($\downarrow$) & \multicolumn{4}{c} {\textbf{\euroauc} ($\uparrow$)} & ($\downarrow$) & \multicolumn{4}{c}{\textbf{\euroauc} ($\uparrow$)}\\
\cmidrule(r){2-6}
\cmidrule(r){7-11}
\cmidrule(r){12-16}
\textbf{estimator} & {\textbf{smECE}} & {low} & {med} & {high} & {all} & {\textbf{smECE}} & {low} & {med} & {high} & {all} & {\textbf{smECE}} & {low} & {med} & {high} & {all}\\
\midrule
Raw Conf & 0.38 & 0.12 & 0.24 & 0.62 & 0.32 & 0.16 & 0.34 & 0.69 & 0.91 & 0.65 & 0.05 & 0.61 & 0.90 & 0.98 & 0.83 \\
HRE & 0.19 & 0.94 & 0.68 & 0.62 & 0.75 & 0.09 & 0.68 & 0.69 & 0.91 & 0.76 & 0.03 & 0.71 & 0.90 & 0.98 & 0.86 \\  
\kernelReg & 0.19 & 0.94 & 0.67 & 0.62 & 0.75 & 0.11 & 0.67 & 0.69 & 0.91 & 0.76 & 0.04 & 0.70 & 0.90 & 0.98 & 0.86 \\
Platt & 0.19 & 0.94 & 0.68 & 0.62 & 0.75 & 0.09 & 0.68 & 0.69 & 0.91 & 0.76 & 0.03 & 0.70 & 0.90 & 0.98 & 0.86 \\
Isotonic & 0.19 & 0.94 & 0.68 & 0.62 & 0.75 & 0.10 & 0.67 & 0.69 & 0.91 & 0.76 & 0.03 & 0.70 & 0.90 & 0.98 & 0.86 \\
Beta & 0.19 & 0.94 & 0.68 & 0.62 & 0.75 & 0.09 & 0.68 & 0.69 & 0.91 & 0.76 & 0.03 & 0.70 & 0.90 & 0.98 & 0.86 \\ 
\midrule 
\methodNameAbbr cosine & 0.19 & 0.94 & 0.67 & 0.62 & 0.75 & 0.09 & 0.68 & 0.69 & 0.91 & 0.76 & 0.05 & 0.70 & 0.90 & 0.98 & 0.86 \\  
\methodNameAbbr pca10 & 0.19 & 0.94 & 0.67 & 0.62 & 0.75 & 0.10 & 0.69 & 0.69 & 0.91 & 0.77 & 0.04 & 0.70 & 0.90 & 0.98 & 0.86 \\  
\methodNameAbbr pca20 & 0.19 & 0.94 & 0.69 & 0.62 & 0.75 & 0.11 & 0.69 & 0.69 & 0.91 & 0.76 & 0.04 & 0.70 & 0.90 & 0.98 & 0.86 \\  
\methodNameAbbr early act & 0.19 & 0.94 & 0.70 & 0.62 & 0.75 & 0.10 & 0.68 & 0.69 & 0.91 & 0.76 & 0.04 & 0.71 & 0.90 & 0.98 & 0.86 \\  
\methodNameAbbr mid act & 0.19 & 0.94 & 0.70 & 0.62 & 0.76 & 0.10 & 0.70 & 0.69 & 0.91 & 0.77 & 0.04 & 0.72 & 0.90 & 0.98 & 0.86 \\  
\methodNameAbbr late act & 0.19 & 0.94 & 0.70 & 0.62 & 0.76 & 0.10 & 0.68 & 0.69 & 0.91 & 0.76 & 0.03 & 0.71 & 0.90 & 0.98 & 0.86 \\ 
\bottomrule
\end{tabular}
}
\caption{%
Results on the SCITLDR dev set evaluated at different Rouge-L thresholds to test threshold sensitivity on Qwen3-14B. Lower smECE is better, while higher \euroauc is better. 
}
\label{tab:scitldr_threshold_results_Qwen3-14B}
\end{table*}

\begin{table*}[t]
\footnotesize
\centering
\newcolumntype{d}{r@{.}l}
\newcommand\mc[1]{\multicolumn{1}{c}{#1}}
{\setlength{\tabcolsep}{3.5pt}
\begin{tabular}{l@{} c c c c c | c c c c c | c c c c c}
\toprule
& \multicolumn{5}{c}{\textbf{Threshold = 0.2}} & \multicolumn{5}{c}{\textbf{Threshold = 0.3}} & \multicolumn{5}{c}{\textbf{Threshold = 0.4}}\\
\midrule
& ($\downarrow$) & \multicolumn{4}{c}{\textbf{\euroauc} ($\uparrow$)} & ($\downarrow$) & \multicolumn{4}{c} {\textbf{\euroauc} ($\uparrow$)} & ($\downarrow$) & \multicolumn{4}{c}{\textbf{\euroauc} ($\uparrow$)}\\
\cmidrule(r){2-6}
\cmidrule(r){7-11}
\cmidrule(r){12-16}
\textbf{estimator} & {\textbf{smECE}} & {low} & {med} & {high} & {all} & {\textbf{smECE}} & {low} & {med} & {high} & {all} & {\textbf{smECE}} & {low} & {med} & {high} & {all}\\
\midrule
Raw Conf & 0.33 & 0.08 & 0.24 & 0.62 & 0.31 & 0.14 & 0.32 & 0.70 & 0.92 & 0.64 & 0.03 & 0.71 & 0.95 & 0.99 & 0.88 \\
HRE & 0.19 & 0.93 & 0.66 & 0.62 & 0.74 & 0.09 & 0.66 & 0.70 & 0.92 & 0.76 & 0.00 & 0.78 & 0.95 & 0.99 & 0.90 \\  
\kernelReg & 0.19 & 0.94 & 0.67 & 0.62 & 0.74 & 0.09 & 0.67 & 0.70 & 0.92 & 0.76 & 0.01 & 0.78 & 0.95 & 0.99 & 0.90 \\
Platt & 0.19 & 0.94 & 0.67 & 0.62 & 0.75 & 0.09 & 0.67 & 0.70 & 0.92 & 0.76 & 0.01 & 0.78 & 0.95 & 0.99 & 0.90 \\
Isotonic & 0.18 & 0.94 & 0.67 & 0.61 & 0.74 & 0.09 & 0.67 & 0.70 & 0.92 & 0.76 & 0.01 & 0.77 & 0.95 & 0.99 & 0.90 \\
Beta & 0.19 & 0.94 & 0.67 & 0.62 & 0.75 & 0.09 & 0.67 & 0.70 & 0.92 & 0.76 & 0.01 & 0.78 & 0.95 & 0.99 & 0.90 \\
\midrule
\methodNameAbbr cosine & 0.19 & 0.94 & 0.64 & 0.62 & 0.74 & 0.09 & 0.67 & 0.70 & 0.92 & 0.76 & 0.02 & 0.77 & 0.95 & 0.99 & 0.90 \\  
\methodNameAbbr pca10 & 0.19 & 0.94 & 0.65 & 0.63 & 0.74 & 0.10 & 0.69 & 0.70 & 0.92 & 0.77 & 0.02 & 0.77 & 0.95 & 0.99 & 0.90 \\  
\methodNameAbbr pca20 & 0.19 & 0.94 & 0.67 & 0.63 & 0.75 & 0.10 & 0.68 & 0.71 & 0.92 & 0.77 & 0.02 & 0.77 & 0.95 & 0.99 & 0.90 \\  
\methodNameAbbr early act & 0.18 & 0.94 & 0.69 & 0.63 & 0.75 & 0.09 & 0.69 & 0.70 & 0.92 & 0.77 & 0.01 & 0.78 & 0.95 & 0.99 & 0.90 \\ 
\methodNameAbbr mid act & 0.18 & 0.94 & 0.69 & 0.63 & 0.75 & 0.09 & 0.69 & 0.71 & 0.92 & 0.77 & 0.01 & 0.78 & 0.95 & 0.99 & 0.90 \\  
\methodNameAbbr late act & 0.18 & 0.94 & 0.68 & 0.63 & 0.75 & 0.09 & 0.68 & 0.70 & 0.92 & 0.77 & 0.01 & 0.77 & 0.95 & 0.99 & 0.90 \\ 
\bottomrule
\end{tabular}
}
\caption{%
Results on the SCITLDR dev set evaluated at different Rouge-L thresholds to test threshold sensitivity on phi-4. Lower smECE is better, while higher \euroauc is better. 
}
\label{tab:scitldr_threshold_results_phi-4}
\end{table*}

\begin{table*}[t]
\footnotesize
\centering
\newcolumntype{d}{r@{.}l}
\newcommand\mc[1]{\multicolumn{1}{c}{#1}}
{\setlength{\tabcolsep}{3.5pt}
\begin{tabular}{l@{} c c c c c | c c c c c | c c c c c}
\toprule
& \multicolumn{5}{c}{\textbf{Threshold = 0.2}} & \multicolumn{5}{c}{\textbf{Threshold = 0.3}} & \multicolumn{5}{c}{\textbf{Threshold = 0.4}}\\
\midrule
& ($\downarrow$) & \multicolumn{4}{c}{\textbf{\euroauc} ($\uparrow$)} & ($\downarrow$) & \multicolumn{4}{c} {\textbf{\euroauc} ($\uparrow$)} & ($\downarrow$) & \multicolumn{4}{c}{\textbf{\euroauc} ($\uparrow$)}\\
\cmidrule(r){2-6}
\cmidrule(r){7-11}
\cmidrule(r){12-16}
\textbf{estimator} & {\textbf{smECE}} & {low} & {med} & {high} & {all} & {\textbf{smECE}} & {low} & {med} & {high} & {all} & {\textbf{smECE}} & {low} & {med} & {high} & {all}\\
\midrule
Raw Conf & 0.28 & 0.12 & 0.39 & 0.76 & 0.42 & 0.10 & 0.42 & 0.80 & 0.95 & 0.72 & 0.03 & 0.69 & 0.94 & 0.99 & 0.87 \\ 
HRE & 0.15 & 0.89 & 0.55 & 0.76 & 0.73 & 0.05 & 0.66 & 0.80 & 0.95 & 0.80 & 0.01 & 0.76 & 0.94 & 0.99 & 0.90 \\  
\kernelReg & 0.15 & 0.89 & 0.55 & 0.76 & 0.73 & 0.04 & 0.66 & 0.80 & 0.95 & 0.80 & 0.01 & 0.76 & 0.94 & 0.99 & 0.90 \\ 
Platt & 0.15 & 0.89 & 0.55 & 0.76 & 0.73 & 0.04 & 0.66 & 0.80 & 0.95 & 0.80 & 0.01 & 0.76 & 0.94 & 0.99 & 0.90 \\ 
Isotonic & 0.15 & 0.89 & 0.55 & 0.76 & 0.73 & 0.05 & 0.66 & 0.80 & 0.95 & 0.80 & 0.01 & 0.76 & 0.94 & 0.99 & 0.90 \\ 
Beta & 0.15 & 0.89 & 0.55 & 0.76 & 0.73 & 0.04 & 0.66 & 0.80 & 0.95 & 0.80 & 0.01 & 0.76 & 0.94 & 0.99 & 0.90 \\  
\midrule
\methodNameAbbr cosine & 0.15 & 0.88 & 0.59 & 0.76 & 0.74 & 0.07 & 0.68 & 0.79 & 0.95 & 0.80 & 0.03 & 0.76 & 0.94 & 0.99 & 0.89 \\  
\methodNameAbbr pca10 & 0.16 & 0.88 & 0.60 & 0.76 & 0.75 & 0.06 & 0.70 & 0.80 & 0.95 & 0.82 & 0.03 & 0.76 & 0.94 & 0.99 & 0.89 \\  
\methodNameAbbr pca20 & 0.15 & 0.89 & 0.61 & 0.76 & 0.76 & 0.06 & 0.70 & 0.80 & 0.95 & 0.82 & 0.03 & 0.76 & 0.94 & 0.99 & 0.89 \\  
\methodNameAbbr early act & 0.16 & 0.89 & 0.60 & 0.76 & 0.75 & 0.05 & 0.70 & 0.80 & 0.95 & 0.82 & 0.02 & 0.77 & 0.94 & 0.99 & 0.90 \\  
\methodNameAbbr mid act & 0.15 & 0.89 & 0.62 & 0.76 & 0.76 & 0.05 & 0.70 & 0.80 & 0.95 & 0.82 & 0.02 & 0.78 & 0.94 & 0.99 & 0.90 \\  
\methodNameAbbr late act & 0.15 & 0.89 & 0.60 & 0.76 & 0.75 & 0.05 & 0.69 & 0.80 & 0.95 & 0.81 & 0.02 & 0.77 & 0.94 & 0.99 & 0.90 \\ 
\bottomrule
\end{tabular}
}
\caption{%
Results on the SCITLDR dev set evaluated at different Rouge-L thresholds to test threshold sensitivity on SmolLM3-3B. Lower smECE is better, while higher \euroauc is better. 
}
\label{tab:scitldr_threshold_results_SmolLM3-3B}
\end{table*}

\begin{table*}[t]
\footnotesize
\centering
\newcolumntype{d}{r@{.}l}
\newcommand\mc[1]{\multicolumn{1}{c}{#1}}
{\setlength{\tabcolsep}{3.5pt}
\begin{tabular}{l@{} c c c c c | c c c c c | c c c c c}
\toprule
& \multicolumn{5}{c}{\textbf{Threshold = 0.2}} & \multicolumn{5}{c}{\textbf{Threshold = 0.3}} & \multicolumn{5}{c}{\textbf{Threshold = 0.4}}\\
\midrule
& ($\downarrow$) & \multicolumn{4}{c}{\textbf{\euroauc} ($\uparrow$)} & ($\downarrow$) & \multicolumn{4}{c} {\textbf{\euroauc} ($\uparrow$)} & ($\downarrow$) & \multicolumn{4}{c}{\textbf{\euroauc} ($\uparrow$)}\\
\cmidrule(r){2-6}
\cmidrule(r){7-11}
\cmidrule(r){12-16}
\textbf{estimator} & {\textbf{smECE}} & {low} & {med} & {high} & {all} & {\textbf{smECE}} & {low} & {med} & {high} & {all} & {\textbf{smECE}} & {low} & {med} & {high} & {all}\\
\midrule
Raw Conf & 0.39 & 0.14 & 0.30 & 0.68 & 0.37 & 0.15 & 0.41 & 0.76 & 0.94 & 0.70 & 0.02 & 0.76 & 0.96 & 0.99 & 0.90 \\
HRE & 0.20 & 0.92 & 0.58 & 0.69 & 0.73 & 0.09 & 0.64 & 0.76 & 0.94 & 0.78 & 0.01 & 0.81 & 0.96 & 0.99 & 0.92 \\
\kernelReg & 0.19 & 0.92 & 0.59 & 0.69 & 0.73 & 0.08 & 0.64 & 0.76 & 0.94 & 0.78 & 0.01 & 0.81 & 0.96 & 0.99 & 0.92 \\
Platt & 0.19 & 0.92 & 0.58 & 0.69 & 0.73 & 0.08 & 0.64 & 0.76 & 0.94 & 0.78 & 0.00 & 0.81 & 0.96 & 0.99 & 0.92 \\
Isotonic & 0.19 & 0.92 & 0.59 & 0.69 & 0.73 & 0.08 & 0.64 & 0.76 & 0.94 & 0.78 & 0.01 & 0.81 & 0.96 & 0.99 & 0.92 \\
Beta & 0.19 & 0.92 & 0.59 & 0.68 & 0.73 & 0.08 & 0.64 & 0.76 & 0.94 & 0.78 & 0.01 & 0.81 & 0.96 & 0.99 & 0.92 \\
\midrule
\methodNameAbbr cosine & 0.20 & 0.92 & 0.58 & 0.69 & 0.73 & 0.09 & 0.63 & 0.76 & 0.94 & 0.78 & 0.03 & 0.77 & 0.96 & 0.99 & 0.91 \\
\methodNameAbbr pca10 & 0.19 & 0.92 & 0.61 & 0.69 & 0.74 & 0.09 & 0.64 & 0.76 & 0.94 & 0.78 & 0.02 & 0.80 & 0.96 & 0.99 & 0.92 \\
\methodNameAbbr pca20 & 0.19 & 0.92 & 0.61 & 0.69 & 0.74 & 0.09 & 0.65 & 0.76 & 0.94 & 0.78 & 0.02 & 0.80 & 0.96 & 0.99 & 0.92 \\
\methodNameAbbr early act & 0.19 & 0.92 & 0.61 & 0.69 & 0.74 & 0.09 & 0.65 & 0.76 & 0.94 & 0.78 & 0.01 & 0.82 & 0.96 & 0.99 & 0.92 \\
\methodNameAbbr mid act & 0.19 & 0.92 & 0.62 & 0.69 & 0.74 & 0.09 & 0.65 & 0.76 & 0.94 & 0.78 & 0.01 & 0.82 & 0.96 & 0.99 & 0.92 \\
\methodNameAbbr late act & 0.19 & 0.92 & 0.61 & 0.69 & 0.74 & 0.09 & 0.65 & 0.76 & 0.94 & 0.78 & 0.01 & 0.81 & 0.96 & 0.99 & 0.92 \\
\bottomrule
\end{tabular}
}
\caption{%
Results on the SCITLDR dev set evaluated at different Rouge-L thresholds to test threshold sensitivity on Qwen3-30B-A3B-Instruct-2507. Lower smECE is better, while higher \euroauc is better. 
}
\label{tab:scitldr_threshold_results_Qwen3-30B-A3B-Instruct-2507}
\end{table*}

\section{Per Model Results}
We show per model results on APIGen and SCITLDR including posthoc calibration of \methodNameAbbr systems in~\cref{tab: apigen_scitldr_gemma3-12b-it,tab: apigen_scitldr_gemma3-4b-it,tab: apigen_scitldr_Qwen3-4B-Instruct-2507,tab: apigen_scitldr_Qwen3-14B,tab: apigen_scitldr_phi-4,tab: apigen_scitldr_SmolLM3-3B,tab: apigen_scitldr_Qwen3-30B-A3B-Instruct-2507_full,tab: apigen_scitldr_averaged}.
Generally, we find that \methodNameAbbr systems outperform baselines and that posthoc calibration lowers calibration error while maintaining or sometimes improving \euroauc for most models.

\begin{table*}[t]
\footnotesize
\centering
\newcolumntype{d}{r@{.}l}
\newcommand\mc[1]{\multicolumn{1}{c}{#1}}
{\setlength{\tabcolsep}{3.5pt}
\begin{tabular}{l@{} c c c c c | c c c c c }
\toprule
& \multicolumn{5}{c}{\textbf{APIGen}} & \multicolumn{5}{c}{\textbf{SCITLDR}}\\
\midrule
& & \multicolumn{4}{c} {\textbf{\euroauc} ($\uparrow$)} & & \multicolumn{4}{c}{\textbf{\euroauc} ($\uparrow$)}\\
\cmidrule(r){3-6}
\cmidrule(r){8-11}
\textbf{estimator} & {\textbf{smECE} ($\downarrow$)} & {low} & {med} & {high} & {all} & {\textbf{smECE} ($\downarrow$)} & {low} & {med} & {high} & {all} \\
\midrule
Raw Conf & 0.213 & 0.246 & 0.528 & 0.827 & 0.531 & 0.191 & 0.324 & 0.671 & 0.908 & 0.631 \\ HRE & 0.019 & 0.846 & 0.612 & 0.832 & 0.764 & 0.095 & 0.690 & 0.671 & 0.908 & 0.756  \\ \kernelReg & 0.039 & 0.831 & 0.595 & 0.834 & 0.754 & 0.100 & 0.687 & 0.671 & 0.908 & 0.754 \\ Platt & 0.037 & 0.826 & 0.607 & 0.834 & 0.756 & 0.096 & 0.685 & 0.671 & 0.908 & 0.754 \\ Isotonic & 0.014 & 0.831 & 0.590 & 0.834 & 0.752 & 0.096 & 0.691 & 0.671 & 0.908 & 0.756 \\ Beta & 0.021 & 0.831 & 0.587 & 0.834 & 0.751 & 0.096 & 0.686 & 0.671 & 0.908 & 0.754 \\ \midrule \methodNameAbbr cosine & 0.052 & 0.859 & 0.705 & 0.833 & 0.800 & 0.100 & 0.688 & 0.668 & 0.908 & 0.754 \\ \methodNameAbbr cosine Platt & 0.046 & 0.859 & 0.702 & 0.839 & 0.801 & 0.111 & 0.682 & 0.670 & 0.908 & 0.753 \\ \methodNameAbbr cosine Isotonic & 0.046 & 0.861 & 0.703 & 0.838 & 0.801 & 0.116 & 0.677 & 0.666 & 0.908 & 0.749 \\ \methodNameAbbr cosine Beta & 0.032 & 0.861 & 0.706 & 0.839 & 0.803 & 0.111 & 0.682 & 0.671 & 0.908 & 0.753 \\ \methodNameAbbr pca10 & 0.032 & 0.912 & 0.820 & 0.879 & 0.871 & 0.108 & 0.698 & 0.676 & 0.908 & 0.760 \\ \methodNameAbbr pca10 Platt & 0.024 & 0.916 & 0.819 & 0.879 & 0.872 & 0.107 & 0.700 & 0.675 & 0.908 & 0.760 \\ \methodNameAbbr pca10 Isotonic & 0.026 & 0.913 & 0.819 & 0.877 & 0.870 & 0.107 & 0.687 & 0.676 & 0.908 & 0.757 \\ \methodNameAbbr pca10 Beta & 0.024 & 0.917 & 0.819 & 0.879 & 0.872 & 0.107 & 0.700 & 0.674 & 0.908 & 0.760 \\ \methodNameAbbr pca20 & 0.048 & 0.912 & 0.825 & 0.871 & 0.870 & 0.113 & 0.702 & 0.674 & 0.908 & 0.761 \\ \methodNameAbbr pca20 Platt & 0.036 & 0.919 & 0.823 & 0.872 & 0.872 & 0.109 & 0.705 & 0.675 & 0.908 & 0.762 \\ \methodNameAbbr pca20 Isotonic & 0.025 & 0.921 & 0.815 & 0.867 & 0.868 & 0.111 & 0.697 & 0.674 & 0.909 & 0.759 \\ \methodNameAbbr pca20 Beta & 0.027 & 0.921 & 0.824 & 0.874 & 0.873 & 0.110 & 0.706 & 0.675 & 0.908 & 0.762 \\ \methodNameAbbr early act & 0.078 & 0.894 & 0.805 & 0.865 & 0.854 & 0.105 & 0.699 & 0.672 & 0.908 & 0.759 \\ \methodNameAbbr early act Platt & 0.036 & 0.909 & 0.806 & 0.872 & 0.863 & 0.112 & 0.698 & 0.672 & 0.908 & 0.759 \\ \methodNameAbbr early act Isotonic & 0.020 & 0.908 & 0.795 & 0.866 & 0.857 & 0.106 & 0.691 & 0.672 & 0.908 & 0.756 \\ \methodNameAbbr early act Beta & 0.032 & 0.911 & 0.806 & 0.872 & 0.863 & 0.112 & 0.698 & 0.672 & 0.908 & 0.759 \\ \methodNameAbbr mid act & 0.086 & 0.895 & 0.826 & 0.873 & 0.864 & 0.103 & 0.705 & 0.672 & 0.908 & 0.761 \\ \methodNameAbbr mid act Platt & 0.031 & 0.913 & 0.832 & 0.885 & 0.877 & 0.113 & 0.701 & 0.677 & 0.908 & 0.761 \\ \methodNameAbbr mid act Isotonic & 0.046 & 0.912 & 0.822 & 0.877 & 0.871 & 0.113 & 0.685 & 0.679 & 0.908 & 0.756 \\ \methodNameAbbr mid act Beta & 0.030 & 0.915 & 0.832 & 0.885 & 0.878 & 0.114 & 0.701 & 0.677 & 0.908 & 0.761 \\ \methodNameAbbr late act & 0.084 & 0.902 & 0.831 & 0.874 & 0.869 & 0.104 & 0.692 & 0.671 & 0.908 & 0.756 \\ \methodNameAbbr late act Platt & 0.035 & 0.919 & 0.836 & 0.882 & 0.880 & 0.110 & 0.686 & 0.671 & 0.908 & 0.754 \\ \methodNameAbbr late act Isotonic & 0.051 & 0.908 & 0.824 & 0.877 & 0.870 & 0.113 & 0.681 & 0.671 & 0.908 & 0.753 \\ \methodNameAbbr late act Beta & 0.026 & 0.920 & 0.836 & 0.884 & 0.880 & 0.110 & 0.686 & 0.671 & 0.908 & 0.754 \\
\bottomrule
\end{tabular}
}
\caption{%
Results on the APIGen test subset (left), and on the SCITLDR dev set (right) on gemma3-12b-it. Lower smECE is better, while higher \euroauc is better. 
}
\label{tab: apigen_scitldr_gemma3-12b-it}
\end{table*}

\begin{table*}[t]
\footnotesize
\centering
\newcolumntype{d}{r@{.}l}
\newcommand\mc[1]{\multicolumn{1}{c}{#1}}
{\setlength{\tabcolsep}{3.5pt}
\begin{tabular}{l@{} c c c c c | c c c c c }
\toprule
& \multicolumn{5}{c}{\textbf{APIGen}} & \multicolumn{5}{c}{\textbf{SCITLDR}}\\
\midrule
& & \multicolumn{4}{c} {\textbf{\euroauc} ($\uparrow$)} & & \multicolumn{4}{c}{\textbf{\euroauc} ($\uparrow$)}\\
\cmidrule(r){3-6}
\cmidrule(r){8-11}
\textbf{estimator} & {\textbf{smECE} ($\downarrow$)} & {low} & {med} & {high} & {all} & {\textbf{smECE} ($\downarrow$)} & {low} & {med} & {high} & {all} \\
\midrule
Raw Conf & 0.206 & 0.260 & 0.550 & 0.855 & 0.552 & 0.165 & 0.356 & 0.710 & 0.922 & 0.660 \\ HRE & 0.032 & 0.771 & 0.624 & 0.878 & 0.758 & 0.080 & 0.666 & 0.710 & 0.922 & 0.765  \\ \kernelReg & 0.039 & 0.776 & 0.619 & 0.878 & 0.758 & 0.075 & 0.672 & 0.710 & 0.922 & 0.767 \\ Platt & 0.031 & 0.779 & 0.627 & 0.878 & 0.761 & 0.074 & 0.673 & 0.710 & 0.922 & 0.768 \\ Isotonic & 0.032 & 0.776 & 0.608 & 0.878 & 0.754 & 0.076 & 0.676 & 0.710 & 0.922 & 0.768 \\ Beta & 0.042 & 0.767 & 0.620 & 0.878 & 0.755 & 0.074 & 0.674 & 0.710 & 0.922 & 0.768 \\ \midrule \methodNameAbbr cosine & 0.034 & 0.791 & 0.678 & 0.879 & 0.783 & 0.082 & 0.654 & 0.708 & 0.922 & 0.760 \\ \methodNameAbbr cosine Platt & 0.038 & 0.793 & 0.676 & 0.878 & 0.783 & 0.036 & 0.687 & 0.710 & 0.922 & 0.772 \\ \methodNameAbbr cosine Isotonic & 0.058 & 0.791 & 0.667 & 0.871 & 0.777 & 0.052 & 0.680 & 0.706 & 0.915 & 0.766 \\ \methodNameAbbr cosine Beta & 0.040 & 0.793 & 0.676 & 0.878 & 0.782 & 0.041 & 0.686 & 0.709 & 0.922 & 0.772 \\ \methodNameAbbr pca10 & 0.043 & 0.825 & 0.726 & 0.883 & 0.812 & 0.088 & 0.683 & 0.715 & 0.922 & 0.773 \\ \methodNameAbbr pca10 Platt & 0.033 & 0.826 & 0.732 & 0.882 & 0.813 & 0.041 & 0.707 & 0.713 & 0.922 & 0.780 \\ \methodNameAbbr pca10 Isotonic & 0.045 & 0.816 & 0.727 & 0.878 & 0.807 & 0.039 & 0.701 & 0.713 & 0.922 & 0.778 \\ \methodNameAbbr pca10 Beta & 0.033 & 0.826 & 0.732 & 0.881 & 0.813 & 0.041 & 0.707 & 0.713 & 0.922 & 0.780 \\ \methodNameAbbr pca20 & 0.088 & 0.850 & 0.772 & 0.886 & 0.836 & 0.091 & 0.690 & 0.714 & 0.922 & 0.774 \\ \methodNameAbbr pca20 Platt & 0.033 & 0.860 & 0.787 & 0.894 & 0.847 & 0.044 & 0.713 & 0.715 & 0.922 & 0.783 \\ \methodNameAbbr pca20 Isotonic & 0.036 & 0.851 & 0.787 & 0.893 & 0.844 & 0.043 & 0.710 & 0.717 & 0.922 & 0.782 \\ \methodNameAbbr pca20 Beta & 0.032 & 0.860 & 0.787 & 0.895 & 0.848 & 0.043 & 0.714 & 0.714 & 0.922 & 0.783 \\ \methodNameAbbr early act & 0.037 & 0.834 & 0.724 & 0.885 & 0.812 & 0.083 & 0.696 & 0.712 & 0.922 & 0.775 \\ \methodNameAbbr early act Platt & 0.035 & 0.841 & 0.728 & 0.882 & 0.817 & 0.048 & 0.712 & 0.712 & 0.922 & 0.781 \\ \methodNameAbbr early act Isotonic & 0.028 & 0.840 & 0.726 & 0.879 & 0.815 & 0.046 & 0.718 & 0.714 & 0.922 & 0.784 \\ \methodNameAbbr early act Beta & 0.034 & 0.841 & 0.728 & 0.883 & 0.817 & 0.048 & 0.713 & 0.712 & 0.922 & 0.782 \\ \methodNameAbbr mid act & 0.063 & 0.840 & 0.750 & 0.883 & 0.824 & 0.081 & 0.690 & 0.713 & 0.922 & 0.774 \\ \methodNameAbbr mid act Platt & 0.035 & 0.851 & 0.758 & 0.890 & 0.833 & 0.050 & 0.708 & 0.717 & 0.922 & 0.781 \\ \methodNameAbbr mid act Isotonic & 0.026 & 0.846 & 0.760 & 0.886 & 0.831 & 0.052 & 0.707 & 0.713 & 0.922 & 0.780 \\ \methodNameAbbr mid act Beta & 0.033 & 0.851 & 0.759 & 0.891 & 0.834 & 0.050 & 0.708 & 0.716 & 0.922 & 0.781 \\ \methodNameAbbr late act & 0.076 & 0.843 & 0.752 & 0.883 & 0.826 & 0.081 & 0.688 & 0.712 & 0.922 & 0.773 \\ \methodNameAbbr late act Platt & 0.033 & 0.855 & 0.763 & 0.889 & 0.836 & 0.042 & 0.705 & 0.714 & 0.922 & 0.780 \\ \methodNameAbbr late act Isotonic & 0.034 & 0.850 & 0.759 & 0.889 & 0.833 & 0.043 & 0.705 & 0.710 & 0.922 & 0.778 \\ \methodNameAbbr late act Beta & 0.027 & 0.855 & 0.765 & 0.890 & 0.837 & 0.042 & 0.706 & 0.713 & 0.922 & 0.780 \\ 
\bottomrule
\end{tabular}
}
\caption{%
Results on the APIGen test subset (left), and on the SCITLDR dev set (right) on gemma3-4b-it. Lower smECE is better, while higher \euroauc is better. 
}
\label{tab: apigen_scitldr_gemma3-4b-it}
\end{table*}

\begin{table*}[t]
\footnotesize
\centering
\newcolumntype{d}{r@{.}l}
\newcommand\mc[1]{\multicolumn{1}{c}{#1}}
{\setlength{\tabcolsep}{3.5pt}
\begin{tabular}{l@{} c c c c c | c c c c c }
\toprule
& \multicolumn{5}{c}{\textbf{APIGen}} & \multicolumn{5}{c}{\textbf{SCITLDR}}\\
\midrule
& & \multicolumn{4}{c} {\textbf{\euroauc} ($\uparrow$)} & & \multicolumn{4}{c}{\textbf{\euroauc} ($\uparrow$)}\\
\cmidrule(r){3-6}
\cmidrule(r){8-11}
\textbf{estimator} & {\textbf{smECE} ($\downarrow$)} & {low} & {med} & {high} & {all} & {\textbf{smECE} ($\downarrow$)} & {low} & {med} & {high} & {all} \\
\midrule
Raw Conf & 0.277 & 0.226 & 0.354 & 0.664 & 0.413 & 0.159 & 0.385 & 0.712 & 0.923 & 0.670 \\ HRE & 0.019 & 0.938 & 0.760 & 0.685 & 0.796 & 0.074 & 0.665 & 0.715 & 0.923 & 0.767  \\ \kernelReg & 0.024 & 0.933 & 0.737 & 0.674 & 0.783 & 0.076 & 0.673 & 0.714 & 0.923 & 0.769 \\ Platt & 0.024 & 0.933 & 0.733 & 0.683 & 0.784 & 0.073 & 0.670 & 0.714 & 0.923 & 0.768 \\ Isotonic & 0.010 & 0.933 & 0.736 & 0.674 & 0.783 & 0.073 & 0.676 & 0.714 & 0.923 & 0.770 \\ Beta & 0.016 & 0.933 & 0.736 & 0.669 & 0.781 & 0.073 & 0.672 & 0.714 & 0.923 & 0.769 \\ \midrule \methodNameAbbr cosine & 0.030 & 0.931 & 0.755 & 0.729 & 0.806 & 0.080 & 0.681 & 0.706 & 0.923 & 0.769 \\ \methodNameAbbr cosine Platt & 0.031 & 0.932 & 0.755 & 0.728 & 0.806 & 0.075 & 0.680 & 0.714 & 0.923 & 0.771 \\ \methodNameAbbr cosine Isotonic & 0.036 & 0.926 & 0.753 & 0.723 & 0.802 & 0.072 & 0.680 & 0.713 & 0.923 & 0.771 \\ \methodNameAbbr cosine Beta & 0.031 & 0.932 & 0.755 & 0.728 & 0.806 & 0.075 & 0.680 & 0.713 & 0.923 & 0.771 \\ \methodNameAbbr pca10 & 0.032 & 0.939 & 0.807 & 0.792 & 0.847 & 0.081 & 0.695 & 0.713 & 0.923 & 0.776 \\ \methodNameAbbr pca10 Platt & 0.036 & 0.940 & 0.806 & 0.792 & 0.847 & 0.082 & 0.691 & 0.713 & 0.923 & 0.775 \\ \methodNameAbbr pca10 Isotonic & 0.047 & 0.931 & 0.805 & 0.791 & 0.843 & 0.087 & 0.688 & 0.712 & 0.923 & 0.773 \\ \methodNameAbbr pca10 Beta & 0.037 & 0.939 & 0.805 & 0.793 & 0.847 & 0.082 & 0.691 & 0.713 & 0.923 & 0.775 \\ \methodNameAbbr pca20 & 0.040 & 0.938 & 0.812 & 0.795 & 0.849 & 0.080 & 0.698 & 0.717 & 0.923 & 0.779 \\ \methodNameAbbr pca20 Platt & 0.031 & 0.941 & 0.815 & 0.795 & 0.851 & 0.081 & 0.698 & 0.717 & 0.923 & 0.779 \\ \methodNameAbbr pca20 Isotonic & 0.049 & 0.936 & 0.814 & 0.791 & 0.848 & 0.079 & 0.700 & 0.719 & 0.923 & 0.780 \\ \methodNameAbbr pca20 Beta & 0.030 & 0.940 & 0.814 & 0.796 & 0.851 & 0.080 & 0.699 & 0.716 & 0.923 & 0.779 \\ \methodNameAbbr early act & 0.028 & 0.935 & 0.790 & 0.775 & 0.834 & 0.078 & 0.687 & 0.713 & 0.923 & 0.773 \\ \methodNameAbbr early act Platt & 0.029 & 0.937 & 0.793 & 0.779 & 0.837 & 0.082 & 0.686 & 0.713 & 0.923 & 0.773 \\ \methodNameAbbr early act Isotonic & 0.034 & 0.936 & 0.790 & 0.776 & 0.835 & 0.080 & 0.680 & 0.714 & 0.923 & 0.771 \\ \methodNameAbbr early act Beta & 0.030 & 0.937 & 0.793 & 0.778 & 0.837 & 0.081 & 0.687 & 0.713 & 0.923 & 0.773 \\ \methodNameAbbr mid act & 0.048 & 0.937 & 0.824 & 0.818 & 0.860 & 0.076 & 0.688 & 0.714 & 0.923 & 0.774 \\ \methodNameAbbr mid act Platt & 0.024 & 0.943 & 0.834 & 0.832 & 0.871 & 0.080 & 0.685 & 0.713 & 0.923 & 0.773 \\ \methodNameAbbr mid act Isotonic & 0.032 & 0.942 & 0.833 & 0.827 & 0.868 & 0.083 & 0.678 & 0.713 & 0.923 & 0.770 \\ \methodNameAbbr mid act Beta & 0.024 & 0.943 & 0.834 & 0.832 & 0.871 & 0.079 & 0.685 & 0.714 & 0.923 & 0.773 \\ \methodNameAbbr late act & 0.055 & 0.937 & 0.821 & 0.811 & 0.857 & 0.076 & 0.681 & 0.714 & 0.923 & 0.772 \\ \methodNameAbbr late act Platt & 0.028 & 0.943 & 0.831 & 0.821 & 0.866 & 0.083 & 0.669 & 0.716 & 0.923 & 0.768 \\ \methodNameAbbr late act Isotonic & 0.023 & 0.943 & 0.832 & 0.821 & 0.866 & 0.085 & 0.661 & 0.717 & 0.923 & 0.766 \\ \methodNameAbbr late act Beta & 0.029 & 0.943 & 0.831 & 0.821 & 0.866 & 0.083 & 0.669 & 0.716 & 0.923 & 0.768 \\
\bottomrule
\end{tabular}
}
\caption{%
Results on the APIGen test subset (left), and on the SCITLDR dev set (right) on Qwen3-4B-Instruct-2507. Lower smECE is better, while higher \euroauc is better. 
}
\label{tab: apigen_scitldr_Qwen3-4B-Instruct-2507}
\end{table*}

\begin{table*}[t]
\footnotesize
\centering
\newcolumntype{d}{r@{.}l}
\newcommand\mc[1]{\multicolumn{1}{c}{#1}}
{\setlength{\tabcolsep}{3.5pt}
\begin{tabular}{l@{} c c c c c | c c c c c }
\toprule
& \multicolumn{5}{c}{\textbf{APIGen}} & \multicolumn{5}{c}{\textbf{SCITLDR}}\\
\midrule
& & \multicolumn{4}{c} {\textbf{\euroauc} ($\uparrow$)} & & \multicolumn{4}{c}{\textbf{\euroauc} ($\uparrow$)}\\
\cmidrule(r){3-6}
\cmidrule(r){8-11}
\textbf{estimator} & {\textbf{smECE} ($\downarrow$)} & {low} & {med} & {high} & {all} & {\textbf{smECE} ($\downarrow$)} & {low} & {med} & {high} & {all} \\
\midrule
Raw Conf & 0.315 & 0.196 & 0.308 & 0.593 & 0.364 & 0.164 & 0.344 & 0.689 & 0.914 & 0.646 \\ HRE & 0.036 & 0.922 & 0.769 & 0.690 & 0.795 & 0.094 & 0.675 & 0.688 & 0.914 & 0.758  \\ \kernelReg & 0.038 & 0.932 & 0.739 & 0.685 & 0.787 & 0.105 & 0.672 & 0.688 & 0.914 & 0.757 \\ Platt & 0.043 & 0.932 & 0.731 & 0.691 & 0.786 & 0.094 & 0.682 & 0.688 & 0.914 & 0.760 \\ Isotonic & 0.028 & 0.932 & 0.733 & 0.680 & 0.783 & 0.095 & 0.673 & 0.688 & 0.914 & 0.758 \\ Beta & 0.025 & 0.932 & 0.733 & 0.673 & 0.781 & 0.094 & 0.679 & 0.688 & 0.914 & 0.760 \\ \midrule \methodNameAbbr cosine & 0.036 & 0.934 & 0.770 & 0.739 & 0.816 & 0.094 & 0.675 & 0.687 & 0.914 & 0.758 \\ \methodNameAbbr cosine Platt & 0.023 & 0.934 & 0.773 & 0.741 & 0.817 & 0.089 & 0.684 & 0.688 & 0.914 & 0.761 \\ \methodNameAbbr cosine Isotonic & 0.026 & 0.932 & 0.763 & 0.737 & 0.812 & 0.094 & 0.679 & 0.685 & 0.914 & 0.758 \\ \methodNameAbbr cosine Beta & 0.022 & 0.934 & 0.773 & 0.742 & 0.817 & 0.089 & 0.684 & 0.688 & 0.914 & 0.761 \\ \methodNameAbbr pca10 & 0.039 & 0.937 & 0.816 & 0.800 & 0.852 & 0.102 & 0.694 & 0.692 & 0.914 & 0.766 \\ \methodNameAbbr pca10 Platt & 0.028 & 0.938 & 0.817 & 0.802 & 0.853 & 0.095 & 0.698 & 0.689 & 0.914 & 0.766 \\ \methodNameAbbr pca10 Isotonic & 0.052 & 0.930 & 0.812 & 0.792 & 0.845 & 0.095 & 0.685 & 0.696 & 0.914 & 0.764 \\ \methodNameAbbr pca10 Beta & 0.029 & 0.938 & 0.817 & 0.800 & 0.853 & 0.095 & 0.697 & 0.688 & 0.914 & 0.766 \\ \methodNameAbbr pca20 & 0.052 & 0.938 & 0.823 & 0.810 & 0.858 & 0.106 & 0.687 & 0.690 & 0.914 & 0.763 \\ \methodNameAbbr pca20 Platt & 0.027 & 0.941 & 0.823 & 0.820 & 0.862 & 0.093 & 0.697 & 0.688 & 0.914 & 0.766 \\ \methodNameAbbr pca20 Isotonic & 0.041 & 0.936 & 0.823 & 0.807 & 0.856 & 0.092 & 0.687 & 0.693 & 0.914 & 0.764 \\ \methodNameAbbr pca20 Beta & 0.031 & 0.941 & 0.823 & 0.817 & 0.861 & 0.093 & 0.696 & 0.688 & 0.914 & 0.765 \\ \methodNameAbbr early act & 0.053 & 0.934 & 0.796 & 0.776 & 0.836 & 0.102 & 0.681 & 0.690 & 0.914 & 0.760 \\ \methodNameAbbr early act Platt & 0.025 & 0.937 & 0.800 & 0.791 & 0.844 & 0.093 & 0.685 & 0.688 & 0.914 & 0.762 \\ \methodNameAbbr early act Isotonic & 0.035 & 0.936 & 0.799 & 0.784 & 0.840 & 0.096 & 0.683 & 0.691 & 0.914 & 0.762 \\ \methodNameAbbr early act Beta & 0.024 & 0.937 & 0.800 & 0.791 & 0.844 & 0.094 & 0.685 & 0.689 & 0.914 & 0.762 \\ \methodNameAbbr mid act & 0.064 & 0.937 & 0.834 & 0.815 & 0.863 & 0.098 & 0.696 & 0.689 & 0.914 & 0.766 \\ \methodNameAbbr mid act Platt & 0.022 & 0.945 & 0.842 & 0.835 & 0.875 & 0.099 & 0.697 & 0.691 & 0.914 & 0.767 \\ \methodNameAbbr mid act Isotonic & 0.023 & 0.943 & 0.839 & 0.836 & 0.873 & 0.103 & 0.692 & 0.693 & 0.914 & 0.765 \\ \methodNameAbbr mid act Beta & 0.023 & 0.944 & 0.842 & 0.835 & 0.874 & 0.099 & 0.698 & 0.691 & 0.914 & 0.767 \\ \methodNameAbbr late act & 0.078 & 0.936 & 0.831 & 0.817 & 0.862 & 0.099 & 0.681 & 0.689 & 0.914 & 0.761 \\ \methodNameAbbr late act Platt & 0.028 & 0.943 & 0.839 & 0.838 & 0.874 & 0.096 & 0.680 & 0.690 & 0.914 & 0.761 \\ \methodNameAbbr late act Isotonic & 0.032 & 0.942 & 0.835 & 0.836 & 0.872 & 0.092 & 0.662 & 0.690 & 0.914 & 0.754 \\ \methodNameAbbr late act Beta & 0.031 & 0.942 & 0.839 & 0.836 & 0.873 & 0.095 & 0.679 & 0.690 & 0.914 & 0.760 \\
\bottomrule
\end{tabular}
}
\caption{%
Results on the APIGen test subset (left), and on the SCITLDR dev set (right) on Qwen3-14B. Lower smECE is better, while higher \euroauc is better. 
}
\label{tab: apigen_scitldr_Qwen3-14B}
\end{table*}

\begin{table*}[t]
\footnotesize
\centering
\newcolumntype{d}{r@{.}l}
\newcommand\mc[1]{\multicolumn{1}{c}{#1}}
{\setlength{\tabcolsep}{3.5pt}
\begin{tabular}{l@{} c c c c c | c c c c c }
\toprule
& \multicolumn{5}{c}{\textbf{APIGen}} & \multicolumn{5}{c}{\textbf{SCITLDR}}\\
\midrule
& & \multicolumn{4}{c} {\textbf{\euroauc} ($\uparrow$)} & & \multicolumn{4}{c}{\textbf{\euroauc} ($\uparrow$)}\\
\cmidrule(r){3-6}
\cmidrule(r){8-11}
\textbf{estimator} & {\textbf{smECE} ($\downarrow$)} & {low} & {med} & {high} & {all} & {\textbf{smECE} ($\downarrow$)} & {low} & {med} & {high} & {all} \\
\midrule
Raw Conf & 0.210 & 0.216 & 0.566 & 0.864 & 0.545 & 0.143 & 0.323 & 0.703 & 0.920 & 0.645 \\ HRE & 0.020 & 0.789 & 0.595 & 0.862 & 0.749 & 0.090 & 0.665 & 0.702 & 0.918 & 0.761  \\ \kernelReg & 0.014 & 0.796 & 0.595 & 0.864 & 0.752 & 0.087 & 0.672 & 0.703 & 0.920 & 0.764 \\ Platt & 0.049 & 0.827 & 0.666 & 0.864 & 0.786 & 0.088 & 0.670 & 0.704 & 0.920 & 0.764 \\ Isotonic & 0.008 & 0.796 & 0.594 & 0.860 & 0.751 & 0.087 & 0.674 & 0.704 & 0.920 & 0.765 \\ Beta & 0.009 & 0.796 & 0.596 & 0.864 & 0.753 & 0.087 & 0.672 & 0.704 & 0.920 & 0.764 \\ \midrule \methodNameAbbr cosine & 0.042 & 0.912 & 0.817 & 0.884 & 0.872 & 0.087 & 0.672 & 0.702 & 0.920 & 0.764 \\ \methodNameAbbr cosine Platt & 0.038 & 0.912 & 0.821 & 0.883 & 0.872 & 0.094 & 0.671 & 0.703 & 0.920 & 0.764 \\ \methodNameAbbr cosine Isotonic & 0.073 & 0.907 & 0.820 & 0.863 & 0.864 & 0.111 & 0.655 & 0.694 & 0.920 & 0.755 \\ \methodNameAbbr cosine Beta & 0.057 & 0.911 & 0.820 & 0.875 & 0.870 & 0.096 & 0.667 & 0.702 & 0.920 & 0.762 \\ \methodNameAbbr pca10 & 0.028 & 0.936 & 0.858 & 0.905 & 0.900 & 0.097 & 0.686 & 0.704 & 0.920 & 0.769 \\ \methodNameAbbr pca10 Platt & 0.031 & 0.937 & 0.860 & 0.904 & 0.901 & 0.095 & 0.688 & 0.705 & 0.920 & 0.770 \\ \methodNameAbbr pca10 Isotonic & 0.031 & 0.935 & 0.860 & 0.902 & 0.899 & 0.089 & 0.674 & 0.704 & 0.920 & 0.765 \\ \methodNameAbbr pca10 Beta & 0.030 & 0.937 & 0.860 & 0.904 & 0.901 & 0.095 & 0.686 & 0.705 & 0.920 & 0.770 \\ \methodNameAbbr pca20 & 0.052 & 0.939 & 0.890 & 0.919 & 0.916 & 0.098 & 0.684 & 0.706 & 0.920 & 0.769 \\ \methodNameAbbr pca20 Platt & 0.024 & 0.948 & 0.891 & 0.921 & 0.921 & 0.090 & 0.688 & 0.706 & 0.920 & 0.771 \\ \methodNameAbbr pca20 Isotonic & 0.026 & 0.939 & 0.891 & 0.922 & 0.918 & 0.086 & 0.676 & 0.704 & 0.920 & 0.766 \\ \methodNameAbbr pca20 Beta & 0.022 & 0.948 & 0.893 & 0.921 & 0.921 & 0.090 & 0.689 & 0.705 & 0.920 & 0.771 \\ \methodNameAbbr early act & 0.052 & 0.934 & 0.877 & 0.908 & 0.904 & 0.092 & 0.686 & 0.705 & 0.920 & 0.770 \\ \methodNameAbbr early act Platt & 0.041 & 0.943 & 0.882 & 0.908 & 0.911 & 0.090 & 0.685 & 0.709 & 0.920 & 0.770 \\ \methodNameAbbr early act Isotonic & 0.033 & 0.946 & 0.880 & 0.908 & 0.912 & 0.093 & 0.681 & 0.706 & 0.919 & 0.768 \\ \methodNameAbbr early act Beta & 0.022 & 0.947 & 0.884 & 0.910 & 0.914 & 0.090 & 0.685 & 0.708 & 0.920 & 0.770 \\ \methodNameAbbr mid act & 0.048 & 0.936 & 0.875 & 0.916 & 0.909 & 0.091 & 0.688 & 0.706 & 0.920 & 0.770 \\ \methodNameAbbr mid act Platt & 0.028 & 0.943 & 0.873 & 0.920 & 0.912 & 0.091 & 0.687 & 0.708 & 0.920 & 0.771 \\ \methodNameAbbr mid act Isotonic & 0.052 & 0.936 & 0.867 & 0.908 & 0.904 & 0.095 & 0.682 & 0.704 & 0.920 & 0.768 \\ \methodNameAbbr mid act Beta & 0.032 & 0.944 & 0.873 & 0.916 & 0.911 & 0.091 & 0.687 & 0.708 & 0.920 & 0.771 \\ \methodNameAbbr late act & 0.042 & 0.936 & 0.875 & 0.916 & 0.909 & 0.091 & 0.680 & 0.705 & 0.920 & 0.767 \\ \methodNameAbbr late act Platt & 0.024 & 0.942 & 0.877 & 0.919 & 0.913 & 0.088 & 0.684 & 0.705 & 0.920 & 0.769 \\ \methodNameAbbr late act Isotonic & 0.040 & 0.936 & 0.879 & 0.906 & 0.907 & 0.090 & 0.676 & 0.703 & 0.920 & 0.765 \\ \methodNameAbbr late act Beta & 0.031 & 0.942 & 0.878 & 0.917 & 0.912 & 0.088 & 0.684 & 0.705 & 0.920 & 0.769 \\
\bottomrule
\end{tabular}
}
\caption{%
Results on the APIGen test subset (left), and on the SCITLDR dev set (right) on phi-4. Lower smECE is better, while higher \euroauc is better. 
}
\label{tab: apigen_scitldr_phi-4}
\end{table*}

\begin{table*}[t]
\footnotesize
\centering
\newcolumntype{d}{r@{.}l}
\newcommand\mc[1]{\multicolumn{1}{c}{#1}}
{\setlength{\tabcolsep}{3.5pt}
\begin{tabular}{l@{} c c c c c | c c c c c }
\toprule
& \multicolumn{5}{c}{\textbf{APIGen}} & \multicolumn{5}{c}{\textbf{SCITLDR}}\\
\midrule
& & \multicolumn{4}{c} {\textbf{\euroauc} ($\uparrow$)} & & \multicolumn{4}{c}{\textbf{\euroauc} ($\uparrow$)}\\
\cmidrule(r){3-6}
\cmidrule(r){8-11}
\textbf{estimator} & {\textbf{smECE} ($\downarrow$)} & {low} & {med} & {high} & {all} & {\textbf{smECE} ($\downarrow$)} & {low} & {med} & {high} & {all} \\
\midrule
Raw Conf & 0.072 & 0.528 & 0.861 & 0.968 & 0.783 & 0.097 & 0.416 & 0.799 & 0.950 & 0.719 \\ HRE & 0.007 & 0.681 & 0.861 & 0.968 & 0.835 & 0.046 & 0.655 & 0.799 & 0.950 & 0.800  \\ \kernelReg & 0.004 & 0.680 & 0.861 & 0.968 & 0.834 & 0.044 & 0.655 & 0.799 & 0.950 & 0.800 \\ Platt & 0.007 & 0.679 & 0.861 & 0.968 & 0.834 & 0.045 & 0.656 & 0.799 & 0.950 & 0.800 \\ Isotonic & 0.006 & 0.694 & 0.861 & 0.968 & 0.839 & 0.046 & 0.655 & 0.799 & 0.950 & 0.800 \\ Beta & 0.016 & 0.682 & 0.861 & 0.968 & 0.835 & 0.045 & 0.655 & 0.799 & 0.950 & 0.800 \\ \midrule \methodNameAbbr cosine & 0.023 & 0.757 & 0.864 & 0.968 & 0.862 & 0.066 & 0.676 & 0.791 & 0.950 & 0.804 \\ \methodNameAbbr cosine Platt & 0.041 & 0.745 & 0.864 & 0.968 & 0.858 & 0.039 & 0.684 & 0.799 & 0.950 & 0.810 \\ \methodNameAbbr cosine Isotonic & 0.035 & 0.750 & 0.863 & 0.968 & 0.859 & 0.039 & 0.690 & 0.799 & 0.950 & 0.812 \\ \methodNameAbbr cosine Beta & 0.041 & 0.745 & 0.864 & 0.968 & 0.858 & 0.038 & 0.687 & 0.799 & 0.950 & 0.811 \\ \methodNameAbbr pca10 & 0.050 & 0.845 & 0.883 & 0.968 & 0.898 & 0.056 & 0.702 & 0.797 & 0.950 & 0.815 \\ \methodNameAbbr pca10 Platt & 0.033 & 0.857 & 0.890 & 0.960 & 0.902 & 0.043 & 0.708 & 0.797 & 0.950 & 0.817 \\ \methodNameAbbr pca10 Isotonic & 0.034 & 0.855 & 0.887 & 0.960 & 0.900 & 0.042 & 0.706 & 0.800 & 0.950 & 0.818 \\ \methodNameAbbr pca10 Beta & 0.034 & 0.855 & 0.889 & 0.962 & 0.901 & 0.041 & 0.709 & 0.798 & 0.950 & 0.818 \\ \methodNameAbbr pca20 & 0.068 & 0.878 & 0.895 & 0.968 & 0.913 & 0.058 & 0.702 & 0.798 & 0.950 & 0.816 \\ \methodNameAbbr pca20 Platt & 0.027 & 0.899 & 0.917 & 0.963 & 0.926 & 0.041 & 0.710 & 0.798 & 0.950 & 0.819 \\ \methodNameAbbr pca20 Isotonic & 0.041 & 0.890 & 0.899 & 0.963 & 0.917 & 0.037 & 0.707 & 0.799 & 0.950 & 0.818 \\ \methodNameAbbr pca20 Beta & 0.028 & 0.898 & 0.916 & 0.964 & 0.926 & 0.039 & 0.712 & 0.798 & 0.950 & 0.819 \\ \methodNameAbbr early act & 0.053 & 0.835 & 0.875 & 0.968 & 0.892 & 0.050 & 0.700 & 0.799 & 0.950 & 0.815 \\ \methodNameAbbr early act Platt & 0.037 & 0.866 & 0.892 & 0.959 & 0.905 & 0.038 & 0.705 & 0.798 & 0.950 & 0.817 \\ \methodNameAbbr early act Isotonic & 0.058 & 0.849 & 0.876 & 0.960 & 0.895 & 0.033 & 0.705 & 0.798 & 0.950 & 0.817 \\ \methodNameAbbr early act Beta & 0.039 & 0.866 & 0.890 & 0.960 & 0.905 & 0.036 & 0.707 & 0.798 & 0.950 & 0.818 \\ \methodNameAbbr mid act & 0.077 & 0.860 & 0.885 & 0.968 & 0.904 & 0.049 & 0.705 & 0.799 & 0.950 & 0.817 \\ \methodNameAbbr mid act Platt & 0.025 & 0.911 & 0.923 & 0.972 & 0.935 & 0.038 & 0.709 & 0.798 & 0.950 & 0.818 \\ \methodNameAbbr mid act Isotonic & 0.045 & 0.908 & 0.898 & 0.970 & 0.925 & 0.031 & 0.712 & 0.799 & 0.950 & 0.819 \\ \methodNameAbbr mid act Beta & 0.025 & 0.910 & 0.922 & 0.972 & 0.934 & 0.037 & 0.710 & 0.798 & 0.950 & 0.818 \\ \methodNameAbbr late act & 0.077 & 0.849 & 0.880 & 0.968 & 0.898 & 0.050 & 0.691 & 0.799 & 0.950 & 0.812 \\ \methodNameAbbr late act Platt & 0.030 & 0.903 & 0.913 & 0.971 & 0.929 & 0.036 & 0.693 & 0.799 & 0.950 & 0.813 \\ \methodNameAbbr late act Isotonic & 0.032 & 0.907 & 0.905 & 0.972 & 0.928 & 0.037 & 0.700 & 0.799 & 0.950 & 0.815 \\ \methodNameAbbr late act Beta & 0.029 & 0.903 & 0.912 & 0.971 & 0.929 & 0.036 & 0.693 & 0.799 & 0.950 & 0.813 \\
\bottomrule
\end{tabular}
}
\caption{%
Results on the APIGen test subset (left), and on the SCITLDR dev set (right) on SmolLM3-3B. Lower smECE is better, while higher \euroauc is better. 
}
\label{tab: apigen_scitldr_SmolLM3-3B}
\end{table*}

\begin{table*}[t]
\footnotesize
\centering
\newcolumntype{d}{r@{.}l}
\newcommand\mc[1]{\multicolumn{1}{c}{#1}}
{\setlength{\tabcolsep}{3.5pt}
\begin{tabular}{l@{} c c c c c | c c c c c }
\toprule
& \multicolumn{5}{c}{\textbf{APIGen}} & \multicolumn{5}{c}{\textbf{SCITLDR}}\\
\midrule
& & \multicolumn{4}{c} {\textbf{\euroauc} ($\uparrow$)} & & \multicolumn{4}{c}{\textbf{\euroauc} ($\uparrow$)}\\
\cmidrule(r){3-6}
\cmidrule(r){8-11}
\textbf{estimator} & {\textbf{smECE} ($\downarrow$)} & {low} & {med} & {high} & {all} & {\textbf{smECE} ($\downarrow$)} & {low} & {med} & {high} & {all} \\
\midrule
Raw Conf & 0.261 & 0.465 & 0.471 & 0.594 & 0.510 & 0.146 & 0.413 & 0.760 & 0.939 & 0.701 \\ HRE & 0.028 & 0.941 & 0.799 & 0.718 & 0.820 & 0.088 & 0.637 & 0.763 & 0.940 & 0.779  \\ \kernelReg & 0.043 & 0.938 & 0.760 & 0.688 & 0.797 & 0.083 & 0.641 & 0.763 & 0.940 & 0.780 \\ Platt & 0.025 & 0.938 & 0.753 & 0.669 & 0.788 & 0.082 & 0.643 & 0.763 & 0.940 & 0.781 \\ Isotonic & 0.021 & 0.938 & 0.753 & 0.688 & 0.795 & 0.082 & 0.637 & 0.763 & 0.940 & 0.778 \\ Beta & 0.036 & 0.938 & 0.753 & 0.679 & 0.791 & 0.083 & 0.641 & 0.761 & 0.940 & 0.779 \\ \midrule \methodNameAbbr cosine & 0.031 & 0.946 & 0.817 & 0.757 & 0.841 & 0.089 & 0.634 & 0.760 & 0.940 & 0.776 \\ \methodNameAbbr cosine Platt & 0.025 & 0.945 & 0.818 & 0.758 & 0.841 & 0.057 & 0.658 & 0.761 & 0.940 & 0.785 \\ \methodNameAbbr cosine Isotonic & 0.029 & 0.946 & 0.817 & 0.749 & 0.838 & 0.063 & 0.650 & 0.763 & 0.940 & 0.783 \\ \methodNameAbbr cosine Beta & 0.031 & 0.945 & 0.819 & 0.753 & 0.840 & 0.056 & 0.658 & 0.762 & 0.940 & 0.785 \\ \methodNameAbbr pca10 & 0.025 & 0.946 & 0.834 & 0.813 & 0.865 & 0.092 & 0.637 & 0.762 & 0.940 & 0.778 \\ \methodNameAbbr pca10 Platt & 0.029 & 0.947 & 0.833 & 0.811 & 0.865 & 0.056 & 0.666 & 0.762 & 0.940 & 0.788 \\ \methodNameAbbr pca10 Isotonic & 0.051 & 0.940 & 0.835 & 0.798 & 0.859 & 0.062 & 0.658 & 0.760 & 0.938 & 0.784 \\ \methodNameAbbr pca10 Beta & 0.037 & 0.945 & 0.832 & 0.809 & 0.863 & 0.060 & 0.665 & 0.761 & 0.940 & 0.787 \\ \methodNameAbbr pca20 & 0.027 & 0.947 & 0.843 & 0.812 & 0.868 & 0.092 & 0.653 & 0.765 & 0.940 & 0.785 \\ \methodNameAbbr pca20 Platt & 0.026 & 0.948 & 0.844 & 0.812 & 0.869 & 0.052 & 0.678 & 0.763 & 0.940 & 0.793 \\ \methodNameAbbr pca20 Isotonic & 0.043 & 0.946 & 0.840 & 0.802 & 0.864 & 0.052 & 0.668 & 0.763 & 0.940 & 0.789 \\ \methodNameAbbr pca20 Beta & 0.036 & 0.947 & 0.844 & 0.806 & 0.866 & 0.052 & 0.678 & 0.763 & 0.940 & 0.793 \\ \methodNameAbbr early act & 0.047 & 0.942 & 0.825 & 0.784 & 0.851 & 0.089 & 0.652 & 0.763 & 0.940 & 0.784 \\ \methodNameAbbr early act Platt & 0.032 & 0.945 & 0.834 & 0.788 & 0.857 & 0.055 & 0.675 & 0.763 & 0.940 & 0.791 \\ \methodNameAbbr early act Isotonic & 0.052 & 0.938 & 0.832 & 0.785 & 0.852 & 0.059 & 0.672 & 0.763 & 0.940 & 0.790 \\ \methodNameAbbr early act Beta & 0.033 & 0.944 & 0.834 & 0.788 & 0.856 & 0.056 & 0.674 & 0.763 & 0.940 & 0.791 \\ \methodNameAbbr mid act & 0.057 & 0.943 & 0.844 & 0.821 & 0.870 & 0.088 & 0.653 & 0.763 & 0.940 & 0.784 \\ \methodNameAbbr mid act Platt & 0.026 & 0.950 & 0.854 & 0.830 & 0.879 & 0.053 & 0.676 & 0.762 & 0.940 & 0.792 \\ \methodNameAbbr mid act Isotonic & 0.037 & 0.945 & 0.852 & 0.816 & 0.872 & 0.056 & 0.680 & 0.763 & 0.936 & 0.792 \\ \methodNameAbbr mid act Beta & 0.037 & 0.948 & 0.854 & 0.825 & 0.876 & 0.053 & 0.676 & 0.763 & 0.940 & 0.792 \\ \methodNameAbbr late act & 0.063 & 0.943 & 0.851 & 0.822 & 0.872 & 0.087 & 0.654 & 0.764 & 0.940 & 0.784 \\ \methodNameAbbr late act Platt & 0.027 & 0.951 & 0.856 & 0.831 & 0.880 & 0.061 & 0.675 & 0.763 & 0.940 & 0.791 \\ \methodNameAbbr late act Isotonic & 0.050 & 0.944 & 0.854 & 0.813 & 0.871 & 0.057 & 0.666 & 0.763 & 0.936 & 0.787 \\ \methodNameAbbr late act Beta & 0.038 & 0.951 & 0.857 & 0.821 & 0.877 & 0.063 & 0.675 & 0.762 & 0.939 & 0.791 \\
\bottomrule
\end{tabular}
}
\caption{%
Results on the APIGen test subset (left), and on the SCITLDR dev set (right) on Qwen3-30B-A3B-Instruct-2507. Lower smECE is better, while higher \euroauc is better. 
}
\label{tab: apigen_scitldr_Qwen3-30B-A3B-Instruct-2507_full}
\end{table*}

\begin{table*}[t]
\footnotesize
\centering
\newcolumntype{d}{r@{.}l}
\newcommand\mc[1]{\multicolumn{1}{c}{#1}}
{\setlength{\tabcolsep}{3.5pt}
\begin{tabular}{l@{} c c c c c | c c c c c }
\toprule
& \multicolumn{5}{c}{\textbf{APIGen}} & \multicolumn{5}{c}{\textbf{SCITLDR}}\\
\midrule
& & \multicolumn{4}{c} {\textbf{\euroauc} ($\uparrow$)} & & \multicolumn{4}{c}{\textbf{\euroauc} ($\uparrow$)}\\
\cmidrule(r){3-6}
\cmidrule(r){8-11}
\textbf{estimator} & {\textbf{smECE} ($\downarrow$)} & {low} & {med} & {high} & {all} & {\textbf{smECE} ($\downarrow$)} & {low} & {med} & {high} & {all} \\
\midrule
Raw Conf & 0.222 & 0.305 & 0.520 & 0.766 & 0.528 & 0.152 & 0.366 & 0.721 & 0.925 & 0.667 \\ HRE & 0.023 & 0.841 & 0.717 & 0.805 & 0.788 & 0.081 & 0.665 & 0.721 & 0.925 & 0.769  \\ \kernelReg & 0.029 & 0.841 & 0.701 & 0.799 & 0.781 & 0.082 & 0.667 & 0.721 & 0.925 & 0.770 \\ Platt & 0.031 & 0.845 & 0.711 & 0.798 & 0.785 & 0.079 & 0.668 & 0.721 & 0.925 & 0.771 \\ Isotonic & 0.017 & 0.843 & 0.696 & 0.797 & 0.780 & 0.079 & 0.669 & 0.721 & 0.925 & 0.771 \\ Beta & 0.024 & 0.840 & 0.698 & 0.795 & 0.778 & 0.079 & 0.668 & 0.721 & 0.925 & 0.771 \\ \midrule \methodNameAbbr cosine & 0.035 & 0.876 & 0.772 & 0.827 & 0.826 & 0.085 & 0.669 & 0.717 & 0.925 & 0.769 \\ \methodNameAbbr cosine Platt & 0.035 & 0.874 & 0.773 & 0.828 & 0.825 & 0.072 & 0.678 & 0.721 & 0.925 & 0.774 \\ \methodNameAbbr cosine Isotonic & 0.043 & 0.873 & 0.769 & 0.821 & 0.822 & 0.078 & 0.673 & 0.718 & 0.924 & 0.771 \\ \methodNameAbbr cosine Beta & 0.036 & 0.874 & 0.773 & 0.826 & 0.825 & 0.073 & 0.678 & 0.721 & 0.925 & 0.774 \\ \methodNameAbbr pca10 & 0.036 & 0.906 & 0.821 & 0.863 & 0.864 & 0.089 & 0.685 & 0.723 & 0.925 & 0.777 \\ \methodNameAbbr pca10 Platt & 0.030 & 0.909 & 0.822 & 0.861 & 0.865 & 0.074 & 0.694 & 0.722 & 0.925 & 0.779 \\ \methodNameAbbr pca10 Isotonic & 0.041 & 0.903 & 0.821 & 0.857 & 0.860 & 0.075 & 0.686 & 0.723 & 0.925 & 0.777 \\ \methodNameAbbr pca10 Beta & 0.032 & 0.908 & 0.822 & 0.861 & 0.864 & 0.074 & 0.694 & 0.722 & 0.925 & 0.779 \\ \methodNameAbbr pca20 & 0.054 & 0.915 & 0.837 & 0.866 & 0.873 & 0.091 & 0.688 & 0.723 & 0.925 & 0.778 \\ \methodNameAbbr pca20 Platt & 0.029 & 0.922 & 0.843 & 0.868 & 0.878 & 0.073 & 0.698 & 0.723 & 0.925 & 0.782 \\ \methodNameAbbr pca20 Isotonic & 0.037 & 0.917 & 0.838 & 0.864 & 0.874 & 0.072 & 0.692 & 0.724 & 0.925 & 0.780 \\ \methodNameAbbr pca20 Beta & 0.030 & 0.922 & 0.843 & 0.868 & 0.878 & 0.072 & 0.699 & 0.723 & 0.925 & 0.782 \\ \methodNameAbbr early act & 0.050 & 0.901 & 0.813 & 0.852 & 0.855 & 0.086 & 0.686 & 0.722 & 0.925 & 0.777 \\ \methodNameAbbr mid act & 0.063 & 0.907 & 0.834 & 0.871 & 0.871 & 0.084 & 0.689 & 0.722 & 0.925 & 0.778 \\ \methodNameAbbr late act & 0.068 & 0.907 & 0.834 & 0.870 & 0.870 & 0.084 & 0.681 & 0.722 & 0.925 & 0.775 \\
\bottomrule
\end{tabular}
}
\caption{%
Results on the APIGen test subset (left), and on the SCITLDR dev set (right) averaged across all 7 LLMs we tested. Lower smECE is better, while higher \euroauc is better. 
}
\label{tab: apigen_scitldr_averaged}
\end{table*}

\section{Additional Limitations}
Our work has several limitations. For multi-token generation settings, many tokens are formatting-based tokens and perhaps should not be fully considered in the joint probability when computing raw confidence, but we found it challenging to pinpoint these, so we defaulted to including all of them. Secondly, for many of these tasks more stochasticity in the generation process may offer stronger performance, especially given that different versions of truncation sampling drastically outperform greedy decoding in most text generation settings. However, for replicability and to remove sources of stochasticity, we elected for greedy decoding. Another limitation could be the requirement to have unfettered access to model internals, although for every model someone, often the model developer and anyone in that sub-organization, have access to model internals.

\end{document}